\documentclass[lettersize,journal]{IEEEtran}
\usepackage{amsmath,amssymb,amsfonts}
\usepackage{algorithmic}
\usepackage{algorithm}
\usepackage{array}
\usepackage[caption=false,font=normalsize,labelfont=sf,textfont=sf]{subfig}
\usepackage{textcomp}
\def\BibTeX{{\rm B\kern-.05em{\sc i\kern-.025em b}\kern-.08em
    T\kern-.1667em\lower.7ex\hbox{E}\kern-.125emX}}
\usepackage{stfloats}
\usepackage{url}
\usepackage{verbatim}
\usepackage{graphicx}
\usepackage{cite}
\usepackage{dsfont}
\usepackage{bbding}
\usepackage{multirow}
\usepackage{bm}
\usepackage{threeparttable}
\usepackage{soul}
\usepackage{color}
\usepackage{booktabs}
\usepackage{makecell}

\begin{document}

\title{Learning to better see the unseen: Broad-Deep Mixed Anti-Forgetting Framework for Incremental Zero-Shot Fault Diagnosis}

\author{Jiancheng~Zhao, Jiaqi~Yue, and Chunhui~Zhao,~\IEEEmembership{Senior~Member,~IEEE}
  \thanks{This work is supported by the National Science Fund for Distinguished Young Scholars (No. 62125306), the Zhejiang Key Research and Development Project (2024C01163).}
  \thanks{Jiancheng Zhao, Jiaqi Yue and Chunhui Zhao are with the College of Control Science and Engineering, Zhejiang University, Hangzhou 310027, China. (Email: chhzhao@zju.edu.cn) }
  \thanks{The corresponding author is Chunhui Zhao.}
  \thanks{This work has been submitted to the IEEE for possible publication. Copyright may be transferred without notice, after which this vesion may no longer be accessible.}
}

\markboth{IEEE TRANSACTIONS ON AUTOMATION SCIENCE AND ENGINEERING}%
{Shell \MakeLowercase{\textit{et al.}}: A Sample Article Using IEEEtran.cls for IEEE Journals}


\maketitle

\begin{abstract}
  Zero-shot fault diagnosis (ZSFD) is capable of identifying unseen faults via predicting fault attributes labeled by human experts. We first recognize the demand of ZSFD to deal with continuous changes in industrial processes, i.e., the model's ability to adapt to new fault categories and attributes while avoiding forgetting the diagnosis ability learned previously.
  To overcome the issue that the existing ZSFD paradigm cannot learn from evolving streams of training data in industrial scenarios, the incremental ZSFD (IZSFD) paradigm is proposed for the first time, which incorporates category increment and attribute increment for both traditional ZSFD and generalized ZSFD paradigms. To achieve IZSFD, we present a broad-deep mixed anti-forgetting framework (BDMAFF) that aims to learn from new fault categories and attributes. To tackle the issue of forgetting, BDMAFF effectively accumulates previously acquired knowledge from two perspectives: features and attribute prototypes. The feature memory is established through a deep generative model that employs anti-forgetting training strategies, ensuring the generation quality of historical categories is supervised and maintained.
  The diagnosis model SEEs the UNSEEN faults with the help of generated samples from the generative model.
  The attribute prototype memory is established through a diagnosis model inspired by the broad learning system. Unlike traditional incremental learning algorithms, BDMAFF introduces a memory-driven iterative update strategy for the diagnosis model, which allows the model to learn new faults and attributes without requiring the storage of all historical training samples.
  The effectiveness of the proposed method is verified by a real hydraulic system and the Tennessee-Eastman benchmark process.
\end{abstract}
\def\abstractname{Note to Practitioners}
\begin{abstract}
  This paper addresses the challenge of zero-shot fault diagnosis in practical industrial scenarios, which is caused by the emergence of new fault categories and descriptions. The objective of this study is to develop an incremental zero-shot industrial fault diagnosis method capable of diagnosing both new and existing faults, whether they are seen or unseen, using existing or new fault descriptions. To achieve this, we propose a broad-deep mixed anti-forgetting framework that enables the model to learn from new fault samples or descriptions while retaining its diagnostic capabilities for historical fault categories. Additionally, we design a memory-driven incremental update strategy to facilitate model updates without the need for storing historical data. Experimental evaluations conducted on a real hydraulic system and the benchmark Tennessee-Eastman process demonstrate the feasibility of our approach. In future research, we plan to explore incremental zero-shot fault diagnosis based on various types of fault descriptions.
\end{abstract}
\def\abstractname{abstract}


\section{Introduction}
\IEEEPARstart{F}{ault} diagnosis plays a crucial role in ensuring the reliable operation of industrial equipment. With advancements in sensor technologies, data-driven fault diagnosis methods have made significant progress in recent years \cite{9066890}. The traditional approach, known as supervised fault diagnosis, treats fault diagnosis as a supervised classification task \cite{7882660, 8384293, 8736782}. This approach needs to collect sufficient samples of all kinds of faults.
However, the collection of labeled fault samples is further complicated by the fact that industrial plants often operate normally for extended periods\cite{9540239}.
Additionally, conducting fault simulation experiments on real industrial equipment is not always feasible due to safety concerns \cite{9881217}. Consequently, practical industrial settings often face an absence of training data for target faults, which restricts the diagnostic performance of supervised fault diagnosis \cite{sun2018deep}.

To address the limitation of supervised fault diagnosis, Feng et al. \cite{feng2020fault} proposed the concept of zero-shot fault diagnosis (ZSFD). In ZSFD, \textit{seen faults} refer to faults with both training samples and fault description labeled by human experts, while \textit{unseen faults} refer to faults without training samples but with fault description. Feng et al. summarized the fault description as attributes, such as the cause, and severity of faults, and then developed a mapping model between samples and attributes for attribute prediction. In this way, they can transfer the mapping model from seen faults to unseen faults and thus classify unseen faults by predicting attributes.
Inspired by this study, a series of ZSFD research have been developed which largely followed Feng et al.’s definition of fault attributes.
Hu et al. \cite{9904860} proposed a semantic-consistent embedding (SCE) method that utilizes a Barlow matrix to assess the consistency between sample embeddings and attributes. This approach aims to learn more efficient embeddings to improve the performance of attribute predictors.
Zhang et al. \cite{zhang2023effective} used CNN to extract features from raw signals and employed a bi-linear function for evaluating the compatibility score between features and attributes to find the highest-ranking unseen fault type.
These methods, based on fault attributes defined by experts, have made significant advancements in diagnosing unseen faults whose fault descriptions exist during the model training phase. Despite these advancements, most of them, such as \cite{9904860, CHEN2023113236, chen2023pyramid, zhang2023effective}, fall under the traditional ZSFD (TZSFD), which is under the assumption that only unseen faults would be tested.

Different from TZSFD, generalized ZSFD (GZSFD) aims to diagnose both seen and unseen faults during the testing phase \cite{Yue2024, mou2023variational}, making it a more challenging task. For the GZSD paradigm, Mou et al. \cite{mou2023variational} designed a variational autoencoder based on distributional semantic embedding and cross-modal reconstruction which extracted latent variables from the input for unseen fault sample generation and then they trained the attribute prediction model in a similar way to that in Feng’s work \cite{feng2020fault} based on both real and generated samples.
Zhuo et al. \cite{zhuo2021auxiliary} proposed a generative adversarial model using fault attributes that utilized attribute vectors as input to generate samples for unseen faults and then trained a supervision classifier based on generated unseen fault samples and existing seen fault samples.
Although the TZSFD and GZSFD paradigms have made significant progress in diagnosing unseen faults that have been described in the model training stage, the problem of \textit{model mismatch} remains unresolved, i.e., the model is designed to fit existing fault categories and attributes during model training, but it is unable to adapt to new fault categories or attributes that emerge after the model training phase.

Model mismatch encompasses two aspects: \textit{category mismatch} and \textit{attribute mismatch}. On one hand, when a new fault category emerges that is different from existing seen and unseen faults, existing TZSFD and GZSFD methods may misclassify it as one of the existing categories \cite{feng2020fault, zhuo2021auxiliary}. Although incremental learning algorithms, like broad learning system \cite{7987745, FENG2020486, 9380770} or regularization-based methods\cite{9349197, 2017Overcoming} can be introduced to update models, they often suffer from catastrophic forgetting, which means the ability to diagnose existing fault categories is compromised. We refer to this problem as the category mismatch.
On the other hand, in ZSFD, insufficient fault attributes may not be effective in distinguishing different faults. The fault attributes are predicted for test samples to determine their category labels. Hence, the diagnosis model includes a fixed number of attribute predictors \cite{feng2020fault, 9904860, mou2023variational}. When new attributes are added by human experts to modify the diagnosis model, a complete retraining process is typically required for new attribute predictors corresponding to the new attributes. However, retraining may not always be feasible due to the lack of long-term storage of historical data or the cost associated with storing data \cite{9062316, Xiang_2019_ICCV}. We refer to this problem as the attribute mismatch.

Actually, in practical industrial scenarios, the emergence of new fault categories or attributes is common. On the one hand, samples of \textit{new seen faults} that are different from the existing seen or unseen faults may be collected, after the initial establishment of the fault diagnosis model. Also, as experts' understanding of faults improves, they can analyze \textit{new unseen faults} that differ from the existing unseen faults based on the operating mechanism of the equipment.
On the other hand, for ZSFD,
experts may supplement \textit{new fault attributes} to more comprehensively describe both seen and unseen categories.
Considering the model mismatch issue of ZSFD, we first propose the paradigm of incremental zero-shot fault diagnosis (IZSFD) for addressing the incremental learning issue in both TZSFD and GZSFD scenarios.
IZSFD consists of two subtasks: category increment and attribute increment.
In the category increment task, during each new learning stage, samples of new seen faults are available for training.
In the attribute increment task, during each new learning stage, new fault attributes are added to the set of existing fault attributes.
The aim of IZSFD is twofold: to enhance the diagnostic ability based on new fault categories or attributes while preserving its ability to identify previously encountered faults without the need to store historical data.

In this paper, the IZSFD paradigm is proposed to fulfill the requirements of incremental learning for fault categories and attributes in practical industrial scenarios, which has been overlooked in existing ZSFD research. To overcome the challenges of model mismatch for IZSFD scenarios, we introduce a broad-deep mixed anti-forgetting framework (BDMAFF), which aims to learn new fault categories and attributes while anti-forgetting. This framework contains feature memory based on a deep generative model and attribute prototype memory of the diagnosis model inspired by the broad learning system. For the generative model, we design the attribute anti-forgetting and feature prototype anti-forgetting training strategies to prevent forgetting historical seen faults.
For the diagnosis model, we propose a memory-driven iterative update strategy for updating the attribute prototype matrix and the memory matrix of the diagnosis model while keeping the memory of historical seen faults with the help of generated samples. Moreover, different from the broad learning system\cite{7987745}, this strategy enables incremental updates of the diagnosis model for new faults and attributes without the need for historical samples.
The main contributions of this paper can be summarized as follows:

\begin{enumerate}
  \item To the best of our knowledge, we first consider the demand of ZSFD to deal with continuous changes in practical industrial scenarios. Therefore, we propose the Increment ZSFD paradigm, which introduces category and attribute increment tasks.
  \item We present a broad-deep mixed anti-forgetting framework to mitigate the forgetting issue of the model mismatch problem for both increment tasks, which leverages the feature memory with anti-forgetting training strategies and the attribute prototype memory.
  \item Unlike existing incremental algorithms designed for data increment, the proposed memory-driven iterative update strategy addresses the specific challenge of attribute increment, where only new class-level fault attribute labels rather than data are available for model updating.
\end{enumerate}

The remainder of the paper is organized as follows: In Section \ref{motivation}, we clarify the motivation and challenges of the IZSFD paradigm. Section \ref{method} presents the proposed broad-deep mixed anti-forgetting framework. In Section \ref{experiments}, we evaluate increment tasks of IZSFD on a real hydraulic system and the Tennessee-Eastman benchmark process. Finally, we conclude the paper in Section \ref{conclusion}.

\section{Motivation}
\label{motivation}

In this section, we aim to provide clarification regarding the motivation behind incremental zero-shot fault diagnosis. Then, we discuss the challenges associated with it.

\subsection{Motivation of Incremental Zero-Shot Fault Diagnosis}
\label{sec:izsfd}
\begin{figure}[!t]
  \centering
  \subfloat[]{\includegraphics[width=0.47\textwidth]{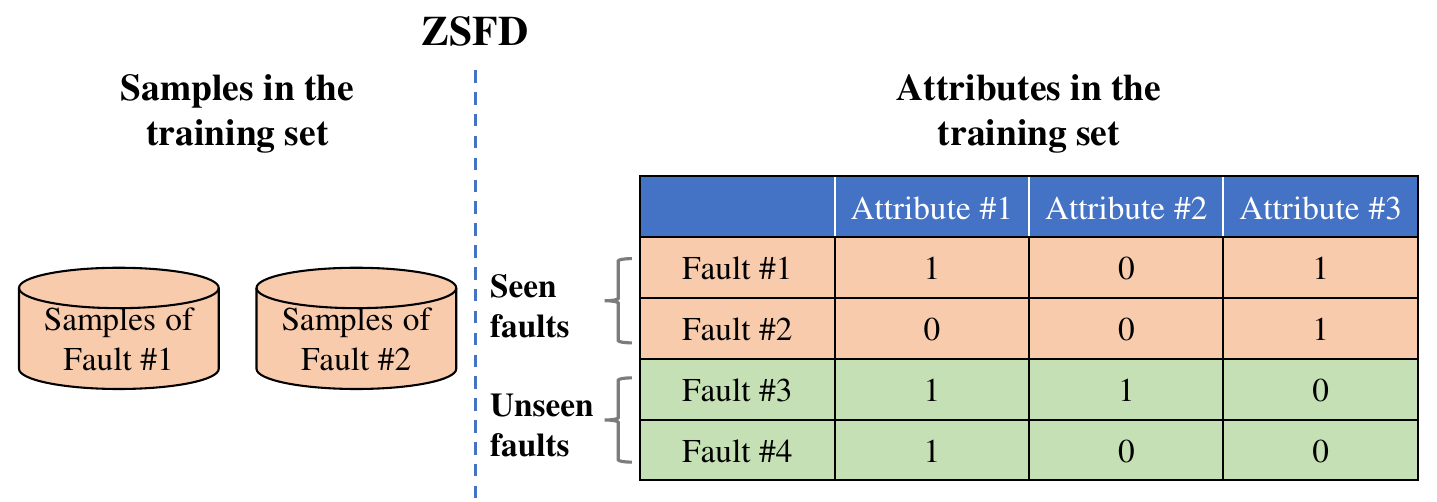}\label{fig:movsub1}}
  \hfill
  \subfloat[]{\includegraphics[width=0.45\textwidth]{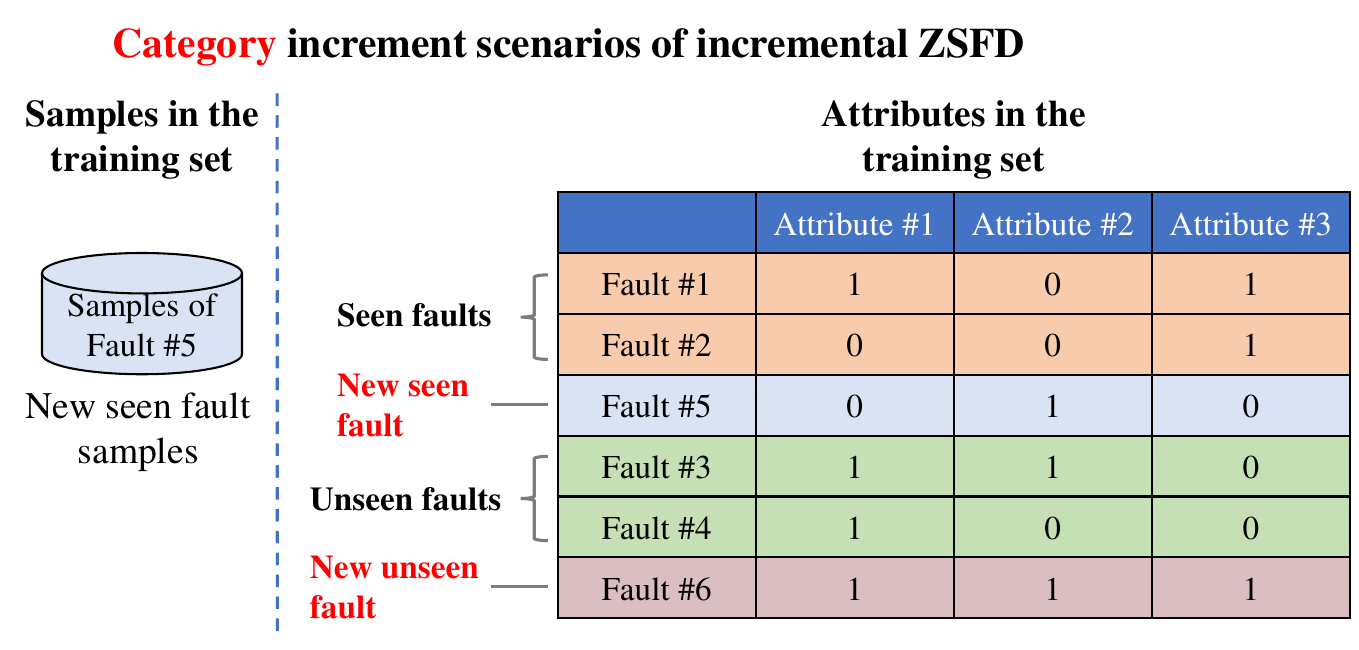}\label{fig:movsub2}}
  \hfill
  \subfloat[]{\includegraphics[width=0.485\textwidth]{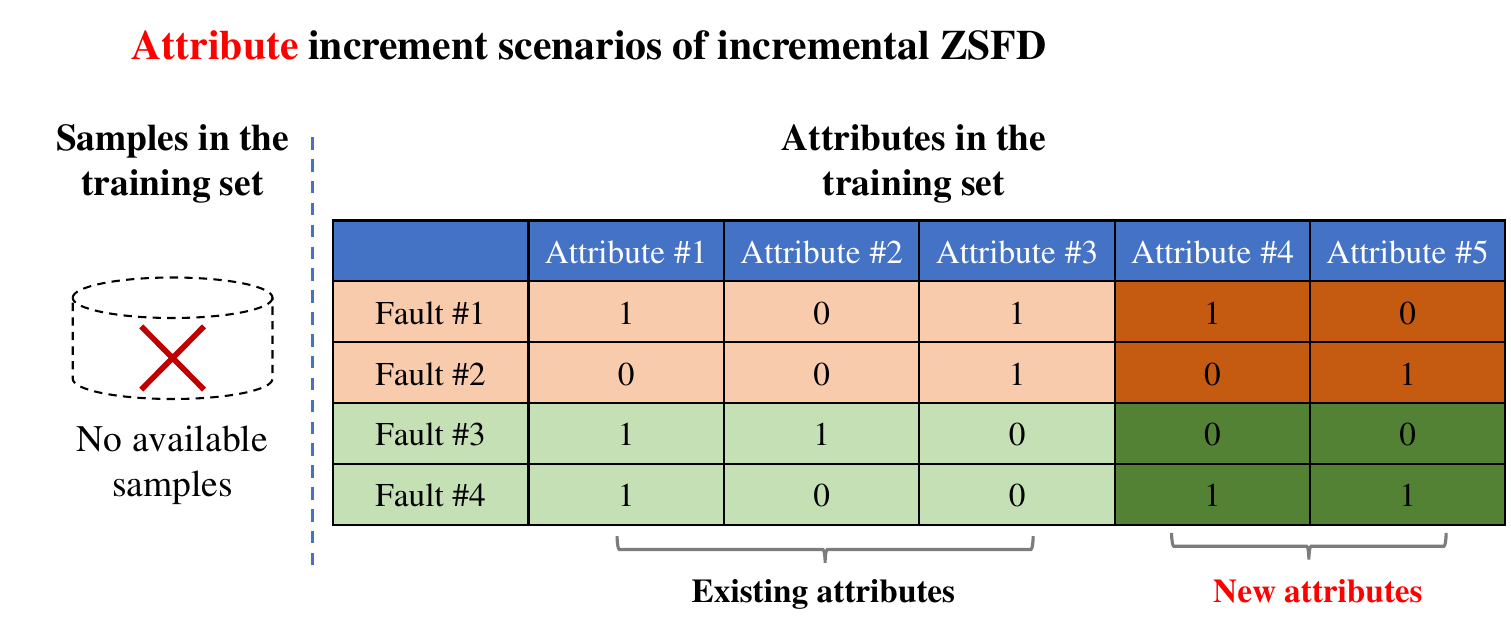}\label{fig:movsub3}}
  \caption{Training set for category and attribute increment tasks. (a) For ZSFD, the training set only contains samples of seen faults (Fault \#1, \#2), along with a fault category-attribute matrix that describes attributes of seen faults (Fault \#1, \#2) and unseen faults (Fault \#3, \#4). (b) For category increment, only samples of new seen faults (Fault \#5) are available for training. Additionally, the category-attribute matrix is expanded to include attributes of new seen faults (Fault \#5) and new unseen faults (Fault \#6). (c) For attribute increment, no training samples are available for model training, while the category-attribute matrix is expanded to include new fault attributes for existing faults.}
  \label{fig:motivation}
\end{figure}

In practical industrial scenarios, the emergence of new fault types is common \cite{9540239, yu2019broad}. Human experts can flexibly learn new fault categories without the need to memorize all historical fault samples. Furthermore, as experience accumulates, human experts gain a deeper understanding of faults, allowing them to analyze different fault types from diverse perspectives. Therefore, to overcome the model mismatch issue mentioned above and simulate the ability of human experts to incrementally learn new fault categories and attributes, we propose the paradigm of incremental zero-shot fault diagnosis for the first time.
IZSFD includes category increment and attribute increment as two subtasks.
To avoid any misunderstandings, we clarify the relevant important concepts related to the fault categories in each learning stage as follows:
\begin{enumerate}
  \item \textit{Seen Faults} refer to fault categories that have had training samples and fault descriptions in previous stages.
  \item \textit{Unseen Faults} refer to fault categories that have had fault descriptions but no samples in previous stages.
  \item \textit{New Faults} refer to fault categories whose corresponding fault descriptions have not appeared in historical stages. New faults contain new seen or unseen faults.
  \item \textit{New Seen Faults} refer to fault categories that appear for the first time in the current learning stage and have both training samples and descriptions.
  \item \textit{New Unseen Faults} refer to fault categories that appear for the first time in the current learning stage and only have descriptions but no training samples.
\end{enumerate}

\begin{table}[!t]
  \centering
  \caption{Testing set of category increment and attribute increment scenarios.}
  \begin{tabular}{p{6em}|p{9em}|p{9em}}
    \toprule
    \multirow{2}{*}{Scenarios}    & \multicolumn{2}{c}{Testing Set}                                                                                                                                                                                                                        \\
    \cmidrule{2-3}                & TZSFD                                                                          & GZSFD                                                                                                                                                                 \\
    \midrule
    ZSFD                          & Unseen faults (Fault \#3, Fault \#4)                                           & Seen faults (Fault \#1, Fault \#2), \newline{} Unseen faults (Fault \#3, Fault \#4)                                                                                   \\
    \midrule
    Category increment scenarios  & Unseen faults (Fault \#3, Fault \#4), \newline{} New unseen faults (Fault \#6) & Seen faults (Fault \#1, Fault \#2), \newline{} Unseen faults (Fault \#3, Fault \#4), \newline{} New seen faults (Fault \#5), \newline{} New unseen faults (Fault \#6) \\
    \midrule
    Attribute increment scenarios & Unseen faults (Fault \#3, Fault \#4)                                           & Seen faults (Fault \#1, Fault \#2), \newline{} Unseen faults (Fault \#3, Fault \#4)                                                                                   \\
    \bottomrule
  \end{tabular}%
  \label{tab:test_set}%
  \begin{tablenotes}
    \item Note: The specific fault categories, i.e., Fault \#1 to Fault \#6, correspond to the fault categories shown in Fig. \ref{fig:motivation}.
  \end{tablenotes}
\end{table}%

For the category increment task, the appearance of new seen faults is due to the collection of samples from new fault types, and the appearance of new unseen faults is due to the deepening cognition of experts, who identify new potential uneen fault categories. The diagnosis model needs to be updated to diagnose new fault categories. The training set for the category increment task is shown in Fig. \ref{fig:movsub2}. In the new learning stage, only samples of new seen faults are available for model training, and the extended fault category-attribute matrix is used to describe all faults. The testing set for the category increment task is shown in Table \ref{tab:test_set}. The goal is to prevent forgetting of historical seen faults, learn new seen faults, and improve the diagnostic ability for all unseen faults. Therefore, the testing set includes unseen faults and new unseen faults (TZSFD), or it includes seen faults, unseen faults, new seen faults, and new unseen faults (GZSFD).

For the attribute increment task, with the addition of fault attributes, both seen and unseen faults can be analyzed and classified from more perspectives, facilitating knowledge transfer from seen categories to unseen categories.
The training set for the attribute increment task is shown in Fig. \ref{fig:movsub3}. In the new learning stage, no real samples are available for model training. Only the extended fault category-attribute matrix can be used to update the diagnosis model. The testing set for the attribute increment task is shown in Table \ref{tab:test_set}. The objective of the attribute increment task is to enhance the diagnostic ability for unseen faults by incorporating newly provided fault attributes from experts, without access to historical samples.

\subsection{The Challenges of Incremental Zero-shot Fault Diagnosis}
\label{sec:challenges}
\textbf{Statement 1}: In the attribute increment task, only new class-level labels are provided, rather than samples. However, existing incremental methods designed for data increment rely on learning from new samples.

Incremental learning tasks typically focus on data increment scenarios, where new training samples are provided in each new learning stage \cite{9349197, 2017Overcoming, 9592832}.
In recent years, exploratory research has been conducted on incremental zero-shot learning in the field of computer vision \cite{9555488, gautam2022tf, kuchibhotla2022unseen}. However, these studies directly apply the data increment setup of incremental learning to zero-shot learning, without considering the auxiliary information of zero-shot learning which is used to identify unseen categories.
Specifically, except for data increment, the auxiliary information used to describe categories, such as attributes, may also increase.
Actually, in real industrial scenarios, experts continuously supplement fault knowledge, leading to an increase in the number of fault attributes to provide a more comprehensive fault description. Therefore, attribute increment is a typical and practically meaningful application scenario for ZSFD. In the attribute increment task, no new training samples are provided, and historical samples are not accessible. Only the dimension of the fault category-attribute matrix is expanded, which is totally different from the data increment task. However, existing incremental learning methods rely on new training samples to update models. Consequently, due to no training samples available, existing increment algorithms are unable to be used for attribute increment \cite{9349197, 2017Overcoming, 2018PackNet, pmlr-v80-serra18a}.

\textbf{Statement 2}: Existing incremental methods focus on remembering historical seen categories for supervised learning scenarios. However, the unseen categories in ZSFD scenarios are ignored by existing incremental methods.

As previously stated, existing incremental learning algorithms focus on preventing the forgetting of seen categories but overlook the challenges posed by unseen categories in ZSFD scenarios. Specifically, regularization-based incremental methods alleviate forgetting by imposing constraints on the model parameters before and after learning new categories. However, these constraints directly limit the parameter space and restrict the model's generalization ability\cite{9349197, 2017Overcoming}. Replay-based methods generate data from historical seen categories during the learning of new categories to prevent forgetting \cite{Xiang_2019_ICCV}. However, relying solely on generating old seen faults may not effectively enhance the diagnostic capability for unseen categories. Parameter isolation methods allocate dedicated parameters for each learning stage, effectively avoiding interference between learning new categories and recalling historical categories \cite{2018PackNet, pmlr-v80-serra18a}. Nevertheless, as the information of seen categories is stored in separate dedicated parameters, the accumulation of knowledge to improve the identification of unseen categories is not achieved.
In sum, conventional incremental learning methods focus on remembering historical categories and lack of special design for zero-shot scenarios. Therefore, they do not consider how to improve the diagnostic capability for unseen faults during the incremental process.

\section{Methodology}
\label{method}
Our objective is to train the proposed model incrementally, considering new seen or unseen fault categories as well as new attributes used to describe faults, without forgetting the existing diagnostic capabilities. This novel fault diagnosis paradigm, which we refer to as IZSFD, is introduced. The specific task setup is described in section \ref{formulation}. We propose a broad-deep mixed anti-forgetting framework by leveraging the feature memory and the fault attribute prototype memory, without the need to store historical data. This approach allows for incremental updates to the model, improving its diagnostic ability for unseen fault categories while ensuring resistance to forgetting previously encountered fault categories. We introduce the overview of the proposed method in section \ref{bd_framework}. In the following sections, we provide a detailed explanation of the update for the generative model and the diagnosis model. In section \ref{sec:incremental}, we present the anti-forgetting incremental strategy for category increment and attribute increment scenarios.

\subsection{Problem Formulation}
\label{formulation}
In ZSFD, the available fault categories for training are represented as $S=\{s_1,\ldots, s_p\}$, where $p$ is the number of seen faults. The fault categories that are not available for training are denoted as $U=\{u_1,\ldots, u_q\}$, where $q$ is the number of unseen faults, and $S \cap U = \emptyset$. The samples belonging to the seen faults are denoted as $\{ \bm{X}^s \in \mathbb{R}^{n^s \times d}, \bm{Y}^s \in \mathbb{R}^{n^s} \}$, where $n^s$ represents the number of training samples, and $d$ denotes the feature dimension. In this research, expert knowledge refers to attributes used to describe faults. The fault attributes of both seen and unseen categories are preprocessed using one-hot coding. After one-hot coding, each attribute is represented as a vector with dimensions equal to the number of possible values for that attribute.
The fault description for category $j$ is denoted as $\bm{a}^j \in \mathbb{R}^{m}$, where $m$ represents the total number of attributes after one-hot coding, $\alpha_i$ denotes the number of possible values for the $i$-th attribute, $\beta$ represents the number of attributes before coding.
The fault category-attributes matrix for all $l$ fault categories is denoted as $\bm{A} = [\bm{A}^s, \bm{A}^u] \in \mathbb{R}^{l \times m}$, where $\bm{A}^s \in \mathbb{R}^{p \times m}$ and $\bm{A}^u \in \mathbb{R}^{q \times m}$ are the attribute matrices for $S$ and $U$, respectively. It is important to note that both the attribute matrices $\bm{A}^s$ and $\bm{A}^u$ are available during the training and testing stages since the descriptions are at the class level rather than the sample level. To incorporate the attribute information, we combine the training labels $\bm{Y}^s$ with the fault category-attributes matrix $\bm{A}^s$ to obtain the training attribute labels $\bm{Z}^s \in \mathbb{R}^{n^s \times m}$. Thus, the training samples from all learning stages can be represented as $\mathbb{D}_{tr}=\{ (\bm{x},y,\bm{z})|\bm{x} \in \bm{X}^s,y \in \bm{Y}^s, \bm{z} \in \bm{Z}^s \}$. Similarly, the testing samples from all learning stages can be denoted as $\mathbb{D}_{ts}=\{ (\bm{x}, y)|\bm{x} \in \bm{X}^t \in \mathbb{R}^{n^t \times d},y \in \bm{Y}^t \in \mathbb{R}^{n^t} \}$, where $\bm{X}^t$ contains test samples from both seen and unseen categories and $\bm{Y}^t$ contains the corresponding target class labels.

\textbf{Settings for the category incremental task:} The seen fault categories $S$ are divided into $T$ non-overlapping parts based on their labels. Similarly, the unseen fault categories $U$ are also divided into $T$ non-overlapping parts based on their labels. During each learning stage, one part of the seen fault categories serves as the training set, while the training samples from the other parts are inaccessible. For testing in learning stage $k$, the test set for seen categories includes all the seen classes encountered from the beginning up to the current stage. The test set for unseen categories consists of all the unseen classes from the first $k$ parts of the unseen categories. In terms of the fault category-attributes matrix, the category incremental task is reflected in the increasing number of rows.

\textbf{Settings for the attribute incremental task:} All $m$ attributes are evenly devided into $T$ non-overlapping parts. In each stage, one part of the attributes is added to the category-attributes matrix, building upon the matrix from the previous stage. The attribute increment task is observed as an increase in the number of columns in the category-attributes matrix.

\subsection{Broad-Deep Mixed Anti-Forgetting Framework}
\label{bd_framework}
\begin{figure*}
  \centering
  \subfloat[]{\includegraphics[width=0.85\textwidth]{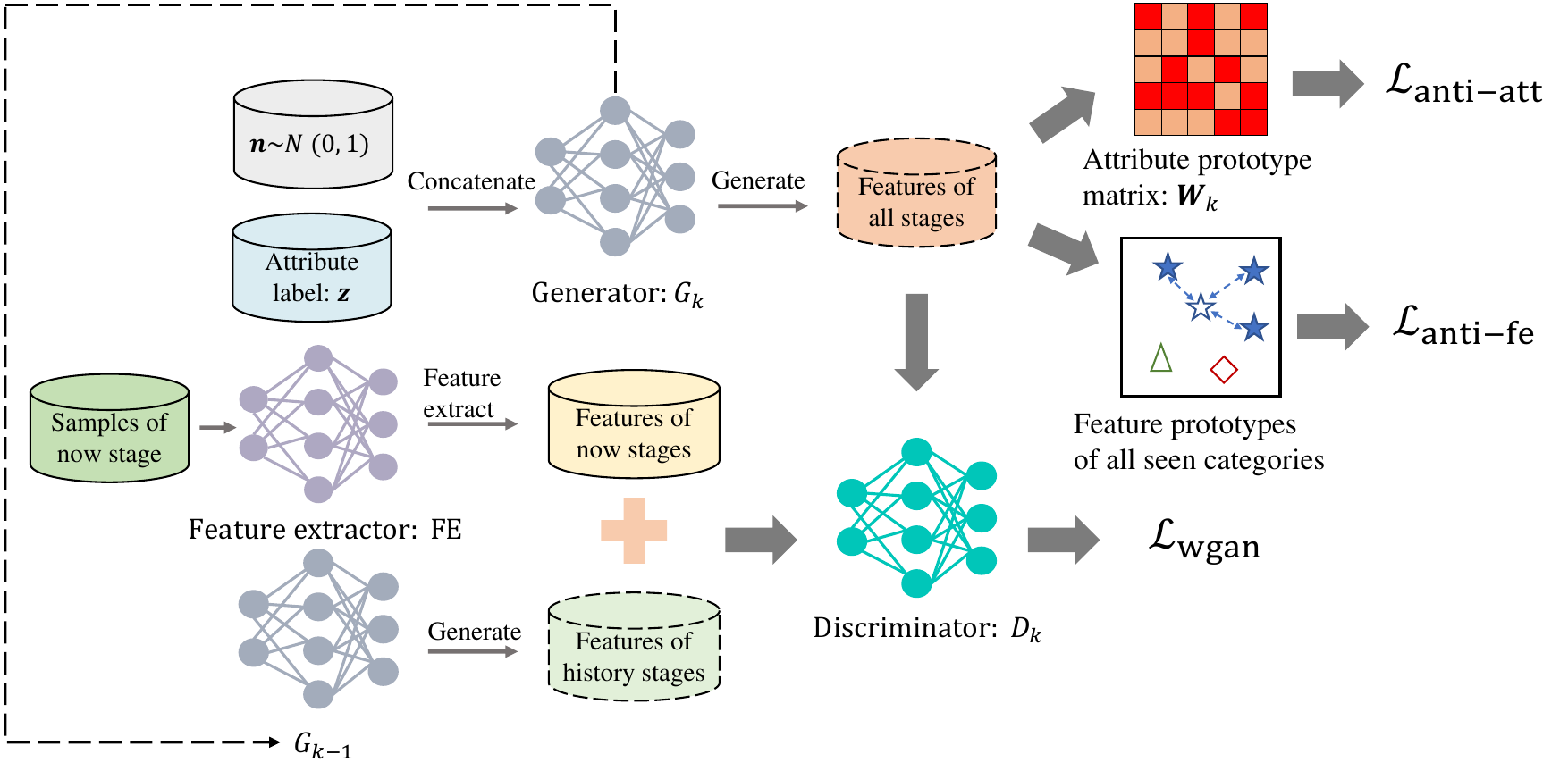}\label{fig:sub1}}
  \hfill
  \subfloat[]{\includegraphics[width=0.85\textwidth]{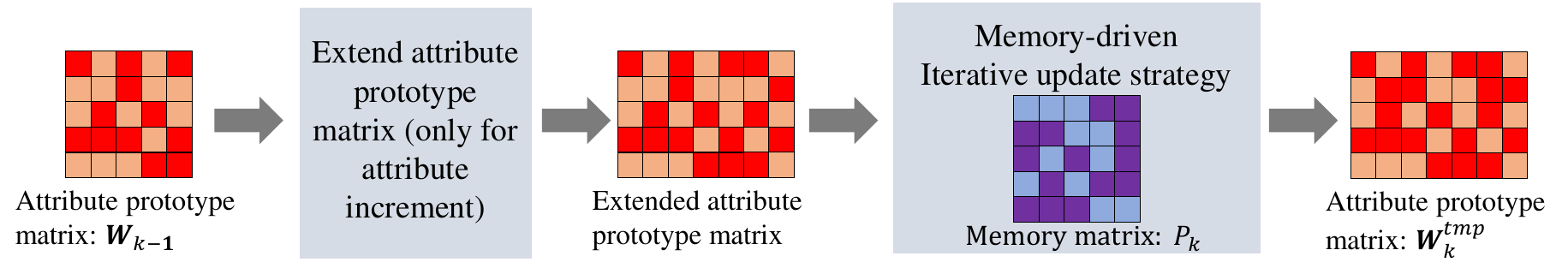}\label{fig:sub2}}
  \caption{Overview of the broad-deep mixed anti-forgetting framework. (a) describe the training process of the generator. In each learning stage, the generator is trained with real features of fault categories corresponding to now stage and generated features of fault categories corresponding to historical stages. Discriminator, the attribute prototype memory and the feature prototypes are used to supervise the generation quality and mitigate the problem of forgetting. (b) describe the update of the attribute prototype matrix used in the diagnosis model. The attribute prototype matrix is extended when new attributes are added to describe faults, and it is updated using a memory-driven iterative update strategy with the help of the memory matrix to keep memory. }
  \label{fig:method}
\end{figure*}


In this section, we present an overview of the proposed BDMAFF for IZSFD. The specific details will be discussed in subsequent sections. Figure \ref{fig:method} illustrates the training process of the proposed BDMAFF method. BDMAFF comprises a generative model ($G$ and $D$) and a diagnosis model ($B$). The generative model generates features of seen faults corresponding to historical learning stage to address forgetting issues and generates features of unseen faults to enhance the diagnosis model's generalization by generating features for unseen faults.
The diagnosis model classifies fault samples by predicting their fault attributes and comparing them with the fault category-attributes matrix. It incorporates a memory matrix to retain attribute prototypes. As the diagnosis model encounters more categories or attributes during incremental learning, its diagnostic capabilities are strengthened. Additionally, the diagnosis model supervises the generation quality of the generative model and mitigates its forgetting issues.

\subsection{The pretraining of the diagnosis model}
\label{sec:pretrain}
In the first increment learning stage, we pre-train the diagnosis model to obtain a feature extractor that transfers samples into more distinguishable features. Based on the pre-trained feature extractor, the generative model generates features rather than samples.
The diagnosis model consists of three components: a feature extractor (FE) comprising a deep neural network, a fault category classification head (CLS) consisting of a single-layer linear layer, and an attribute prototype matrix ($\bm{W}$). The feature extractor's parameters are denoted as $\theta$, the CLS parameters are denoted as $\phi$, and the attribute prototype matrix parameters are denoted as $\bm{W} \in \mathbb{R}^{\hat{d} \times l}$, where $\hat{d}$ represents the dimension of the features extracted by the feature extractor. The CLS maps the extracted features to their respective fault category labels, while $\bm{W}$ maps the features to their corresponding fault attribute labels. To train an effective feature extractor, we initially use the training data from the first stage, denoted as $\mathbb{D}{tr}^1 = \{ (\bm{x},y,\bm{z})|\bm{x} \in \bm{X}^{s}_{1},y \in \bm{Y}^{s}_{1}, \bm{z} \in \bm{Z}^{s}_{1} \}$, to train the feature extractor. The training objective is to minimize the loss function $\mathcal{L}{pre}$:
\begin{equation}
  \mathcal{L}{cls} = \mathcal{L}_{CE}({\rm CLS}(\bm{x}_{fe})), \bm{x}_{fe} = {\rm FE}(\bm{x})
\end{equation}
\begin{equation}
  \mathcal{L}{att} = \sum_{i=1}^{\beta} \mathcal{L}_{NLL}({\rm SoftMax}(\hat{\bm{z}}^i), \bm{z}^i), \hat{\bm{{z}}} = \bm{x}_{fe} \bm{W},
\end{equation}
\begin{equation}
  \mathcal{L}{pre} = \mathcal{L}{cls} + \mathcal{L}{att},
\end{equation}
where $\mathcal{L}_{CE}$ and $\mathcal{L}_{NLL}$ represent the cross-entropy loss and negative log-likelihood loss, respectively. $\hat{\bm{z}}^i$ and $\bm{z}^i$ correspond to subsets of $\hat{\bm{{z}}}$ or $\bm{z}$ that correspond to the $i$-th attribute before one-hot encoding, respectively. After pretraining, we get the updated feature extractor and attribute prototype matrix. We freeze the feature extractor's parameters and only update the attribute prototype matrix in the subsequent learning stages.

\subsection{The update of the generative model}
\label{sec:generative}
We propose a generative model inspired by WGAN-GP \cite{NIPS2017_892c3b1c}, to generate features for both historical seen categories and unseen fault categories. Then the generated features are used for updating the diagnosis model in a new learning stage. By generating features for historical categories, we aim to alleviate the issue of forgetting and accumulate diagnostic capabilities. Furthermore, this enables us to transfer the mapping relationship between fault attributes and fault features to unseen fault categories through the generation of features for unseen fault categories. The generative model consists of a generator ($G$) and a discriminator ($D$). We adapt the objective function of WGAN-GP, which is defined as follows:
\begin{equation}
  \label{equ:wgan}
  \begin{aligned}
    \mathcal{L}_{\mathrm{wgan}}= & \mathbb{E}[D(\bm{x}_{fe}, \bm{z})]-\mathbb{E}[D(\widetilde{\bm{x}}_{fe}, \bm{z})]                                                  \\
                                 & -\lambda \mathbb{E}\left[\left(\left\|\nabla_{\widehat{\bm{x}}_{fe}} D(\widehat{\bm{x}}_{fe}, \bm{z})\right\|_2-1\right)^2\right],
  \end{aligned}
\end{equation}
where $\tilde{\bm{x}}_{fe}=G(\bm{n}, \bm{z})$, and $\widehat{\bm{x}}_{fe}=\gamma \bm{x}_{fe}+(1-\gamma) \widetilde{\bm{x}}_{fe}$, where $\gamma \sim U(0,1)$. Here, $\bm{n}$ represents the noise vector, $\bm{z}$ denotes the fault attribute label, and $\lambda$ is the gradient penalty coefficient. The first two terms in Equation (\ref{equ:wgan}) approximate the Wasserstein distance, while the third term is the gradient penalty that regularizes the gradient of $D$ along the straight line between pairs of real and generated features, ensuring it has a unit norm. Although WGAN-GP is a powerful generative model, it is not directly suitable for our task due to the issue of forgetting historical categories as new categories emerge.

To address the problem of forgetting, we introduce three anti-forgetting strategies. The first strategy is replaying historical faults through sample generation. As shown in Fig. \ref{fig:method}, in each new learning stage, we utilize the previously trained generative model to generate features belonging to categories from the historical stage, which are then used to train the generative model for the current stage. However, relying solely on the replay strategy may be inadequate as the generated feature distribution may not perfectly match the true distribution of historical categories. Over multiple learning stages, the accumulation of generation errors for historical categories can degrade the quality of generated features, leading to forgetting. To mitigate this, we introduce two additional loss terms: the attribute anti-forgetting loss ($\mathcal{L}_{{\mathrm{anti-att}}}$) and the feature prototype anti-forgetting loss ($\mathcal{L}_{\mathrm{anti-fe}}$). The attribute anti-forgetting loss is derived from the attribute prototype matrix of the diagnosis model, which supervises the generation quality of the generative model by assessing whether the generated features contain the correct attribute information. The attribute anti-forgetting loss can be formulated as follows:
\begin{equation}
  \mathcal{L}_{\mathrm{{anti-att}}} = \sum_{i=1}^{\beta} \mathcal{L}_{NLL}({\rm SoftMax}(\tilde{\bm{{z}}}^i), \bm{z}^i), \tilde{\bm{{z}}} = \tilde{\bm{x}}_{fe} W^{ATT}.
\end{equation}

Furthermore, to tackle the issue of forgetting within the generative model itself caused by the accumulation of generation errors, we introduce a feature prototype anti-forgetting loss ($\mathcal{L}_{\mathrm{anti-fe}}$). For each observed fault category, we maintain its associated feature prototype, which represents the mean of the features belonging to that category. In each new learning stage, the generative model is utilized to generate features for all historical fault categories. Subsequently, we calculate the Euclidean distance between the generated features and their corresponding feature prototypes. We define the portion of distances exceeding the threshold $\alpha_{limit}$ as the feature prototype anti-forgetting loss. By incorporating this loss, our objective is to mitigate the problem of distribution shift in the generated feature space for historical categories as the number of learning stages increases. The feature prototype anti-forgetting loss can be formulated as follows:
\begin{equation}
  \mathcal{L}_{\mathrm{anti-fe}} = \mathbb{E}\left [  \sum_{i=1}^{p^k} \left [ \left \| \tilde{\bm{x}}_{fe}^i - \frac{1}{n^i} \sum_{j=1}^{n^i} \bm{x}_{fe}^{i, j} \right \| - \alpha_{limit} \right ]\right ]_+,
\end{equation}
where $n^i$ represents the number of samples in seen category $i$, and $p^k$ denotes the total number of seen categories up to the current learning stage. The complete objective function can be expressed as follows:
\begin{equation}
  \mathcal{L} = \mathcal{L}_{\mathrm{wgan}} + \mathcal{L}_{\mathrm{{anti-att}}} + \mathcal{L}_{\mathrm{anti-fe}},
\end{equation}

\subsection{The update of the diagnosis model}
\label{sec:attribute}
In section \ref{sec:pretrain}, we mention that the pre-trained feature extractor is frozen, and only the attribute prototype matrix needs to be updated in the fault diagnosis model during the new learning stage. Except for the pre-trained feature extractor, the diagnosis model consists of an attribute prototype matrix and a memory matrix. The attribute prototype matrix is used to map the features to their corresponding fault attribute labels. The memory matrix is used to alleviate the problem of forgetting while model updating. During the updating process of the attribute prototype matrix, the samples generated by the generative model are utilized to alleviate forgetting and make the diagnosis model see the unseen faults. In this section, we discuss the updating process of the attribute prototype matrix for both category incremental and attribute incremental tasks, to address the problem of model mismatch.

To tackle the issue of category increment, we draw inspiration from the data increment algorithm of the broad learning system \cite{7987745}. However, the broad learning system's data increment algorithm requires storing all historical training data when updating the model. Considering storage costs and privacy concerns, we aim to avoid storing all the real historical data. Therefore, we introduce a memory-driven iterative update strategy. Specifically, to preserve the memory capacity of fault attribute prototypes and mitigate the problem of forgetting, we introduce a memory matrix. Since we don't have access to the training samples from previous stages, we update the attribute prototype matrix and the memory matrix iteratively. In the initial learning stage, we first initialize the fault attribute prototype matrix $\bm{W}_0$ during the pretraining of the diagnosis model. Then, we initialize the memory matrix $\bm{P}_0$ based on the training samples using the following procedure:
\begin{equation}
  \label{equ:memory}
  \boldsymbol{P}_0=\left(\boldsymbol{X}_{fe, 0}^{\mathrm{T}} \boldsymbol{X}_{fe, 0}\right)^{\dagger}, \boldsymbol{X}_{fe, 0}={\rm FE}(\boldsymbol{X}_{0}),
\end{equation}

During the incremental learning process of our model, the new training samples are represented as $\{\boldsymbol{X}_{fe, k}, \boldsymbol{Z}_{k}\}$. The memory matrix $\boldsymbol{P}_{k}$ and the fault attribute prototype matrix $\bm{W}_{k}$ in learning stage $k$ can be updated as follows:
\begin{equation}
  \label{equ:memory_update}
  \bm{v} = \boldsymbol{P}_{k-1} \boldsymbol{X}_{fe, k}^{\mathrm{T}}\left(\boldsymbol{I}+\boldsymbol{X}_{fe, k} \boldsymbol{P}_{k-1} \boldsymbol{X}_{fe, k}^{\mathrm{T}}\right)^{\dagger}
\end{equation}
\begin{equation}
  \label{equ:memory_update2}
  \boldsymbol{P}_k=\boldsymbol{P}_{k-1}- \bm{v}\boldsymbol{X}_{fe, k} \boldsymbol{P}_{k-1}
\end{equation}
\begin{equation}
  \label{equ:W-ATT_update}
  \bm{W}_{k}=\bm{W}_{k-1}+\boldsymbol{P}_k \boldsymbol{X}_{fe, k}^{\mathrm{T}}\left(\boldsymbol{Z}_{fe, k}-\boldsymbol{X}_{fe, k} \bm{W}_{k-1}\right)
\end{equation}

So the attribute prototype matrix is updated without the need to store historical training samples. Only a memory matrix $\boldsymbol{P} \in \mathbb{R}^{\hat{d} \times \hat{d}}$ is required to maintain the memory, and there is no need to expand the size of $\boldsymbol{P}$ when new samples emerge.

In the attribute increment task, the fault attribute matrix of learning stage $k-1$ is denoted as $\bm{A}_{k-1}$. Consequently, the fault attribute matrix of learning stage $k$ can be expressed as $\bm{A}_{k} = [\bm{A}_{k-1} | \bm{A}_{k}^{new}] \in \mathbb{R}^{l \times m^{k}}$, where $\bm{A}_{k}^{new}$ represents the newly added attribute matrix in stage $k$. In the diagnosis model, the attribute prototype matrix $\bm{W}$ is employed to identify the attributes of input samples. The predicted attributes are then compared with the fault attribute matrix $\bm{A}_{k}$ to determine the fault category. To update the attribute prototype matrix for the attribute increment task, given the unavailability of training data from historical stages, we utilize the generative model to generate features ($\tilde{\bm{x}}_{fe}$) for the seen categories. The attribute prototype matrix of learning stage $k$ is denoted as $\bm{W}_{k} = [\bm{W}_{k-1} | \bm{W}^{new}_{k}]$, where $\bm{W}^{new}_{k}$ represents the newly added portion in stage $k$. The new attribute prototype matrix $\bm{W}_{k}^{new}$ is obtained by minimizing the following loss function:
\begin{equation}
  \mathcal{L}_{\rm seen} = \sum_{i=1}^{\beta_{new}} \mathcal{L}_{NLL}({\rm SoftMax}(\tilde{\bm{z}}^i), \bm{z}^i), \tilde{\bm{{z}}} = \tilde{\bm{x}}_{fe} \times \bm{W}^{new}_{k}.
\end{equation}
where $\beta_{new}$ represents the number of newly added attributes in the $k$th learning stage, and $\tilde{\bm{z}}^i$ and $\bm{z}^i$ correspond to the parts of $\tilde{\bm{{z}}}$ or $\bm{z}$ that correspond to the $i$-th new attribute. Subsequently, to enhance the diagnostic capability for unseen categories, we utilize the generative model to generate features for unseen categories. Then, the attribute prototype matrix $\bm{W}_{k}$ is updated based on equations (\ref{equ:memory_update}), (\ref{equ:memory_update2}), (\ref{equ:W-ATT_update}), utilizing the generated features of unseen faults.

\subsection{Anti-forgetting incremental learning strategy}
\label{sec:incremental}
In this section, we outline the updating process of the proposed BDMAFF model for both category increment and attribute increment tasks. The training process of the proposed BDMAFF model for the category increment task is presented in Algorithm \ref{alg:category}. In each learning stage, we initially pre-train the diagnosis model to acquire the diagnosis ability on new categories. Subsequently, we replay historical seen faults and employ both new and generated historical seen faults to train the generative model. The generative model learns to generate new faults while retaining its memory of historical faults. After that, we utilize the generative model to generate features for unseen faults, to enhance the diagnostic capability for unseen faults. Then we use the generated features of seen and unseen faults to update the attribute prototype matrix again, resulting in $\bm{W}^{tmp}_k$. In the model test process, for each test sample $\bm{x}$, its predicted fault category $\hat{y}$ is determined using $\bm{W}^{tmp}_k$, which can be denoted as follows:
\begin{equation}
  \label{equ:predict}
  \hat{y} = \mathop{\arg\max}\limits_{y} \bm{x}_{fe} \bm{W}^{tmp}_k \bm{A}_k^{\mathrm{T}}.
\end{equation}

\begin{algorithm}[!t]
  \renewcommand{\algorithmicrequire}{\textbf{Input:}}
  \renewcommand{\algorithmicensure}{\textbf{Output:}}
  \caption{Anti-forgetting strategy for category increment}
  \label{alg:category}
  \begin{algorithmic}[1]
    \REQUIRE The training samples of each leaning stage $\{ \bm{x} \in \bm{X}_k^s, y \in \bm{Y}_k^s, \bm{z} \in \bm{Z}_k^s\}$.
    \ENSURE FE: $\theta$, $\bm{W}^{tmp}_k$.
    \FOR{each learning stage $k$}
    \STATE $\slash *$  Diagnosis model pretraining  $* \slash$
    \IF{$k==1$}
    \STATE $\mathcal{L}{pre} = \mathcal{L}{cls} + \mathcal{L}{att}$;
    \STATE Update $\theta$, $\bm{W}_k$ and $\phi$ by minimizing $\mathcal{L}{pre}$;
    \ELSE
    \STATE Update $\bm{W}_k$, $\bm{P}_k$ based on equations (\ref{equ:memory_update}), (\ref{equ:memory_update2}), (\ref{equ:W-ATT_update}) using $\{\bm{X}_k^s, \bm{Z}_k^s\}$;
    \ENDIF
    \STATE $\slash *$  Generative model training  $* \slash$
    \STATE Generate features $\{\tilde{\boldsymbol{X}}_{fe}^s, \boldsymbol{Z}^s\}$ for all history seen categories of stage $k$ using the generative model $G$;
    \STATE $\mathcal{L} = \mathcal{L}_{\mathrm{wgan}} + \lambda_{att} \times \mathcal{L}_{\mathrm{{anti-att}}} + \lambda_{fe} \times \mathcal{L}_{\mathrm{anti-fe}}$;
    \STATE Update $G$ and $D$ by minimizing $\mathcal{L}$ based on  $\{\tilde{\boldsymbol{X}}_{fe}^s, \boldsymbol{Z}^s\}$ and $\{\bm{X}_k^s, \bm{Z}_k^s\}$;
    \STATE $\slash *$  Generate and replay features  $* \slash$
    \STATE Generate features $\{\tilde{\boldsymbol{X}}_{fe}^u, \boldsymbol{Z}^u\}$ for unseen categories of stage $k$ using the generative model $G$;
    \STATE $\slash *$  Diagnosis model training  $* \slash$
    \IF{$k==1$}
    \STATE Initialize $\bm{P}_0$ based on equation (\ref{equ:memory});
    \ENDIF
    \STATE Initialize $\bm{W}^{tmp}_k$ based on $\bm{W}_{k}$;
    \STATE Update  $\bm{W}^{tmp}_k$ based on equations (\ref{equ:memory_update}), (\ref{equ:memory_update2}), (\ref{equ:W-ATT_update}) using generated features of seen and unseen faults, i.e., $\{\tilde{\boldsymbol{X}}_{fe}^u, \boldsymbol{Z}^u\}$, $\{\tilde{\boldsymbol{X}}_{fe}^s, \boldsymbol{Z}^s\}$;
    \ENDFOR
  \end{algorithmic}
\end{algorithm}

\begin{algorithm}[!t]
  \renewcommand{\algorithmicrequire}{\textbf{Input:}}
  \renewcommand{\algorithmicensure}{\textbf{Output:}}
  \caption{Anti-forgetting strategy for attribute increment}
  \label{alg:knowledge}
  \begin{algorithmic}[1]
    \REQUIRE The training samples of the first leaning stage $\{ \bm{x} \in \bm{X}_0^s, y \in \bm{Y}_0^s, \bm{z} \in \bm{Z}_0^s\}$. The fault attribute matrix $\bm{A}_k$ of each learning stage.
    \ENSURE FE: $\theta$, $\bm{W}^{tmp}_k$.
    \FOR{each learning stage $k$}
    \STATE $\slash *$  Diagnosis model pretraining  $* \slash$
    \IF{$k==1$}
    \STATE $\mathcal{L}{pre} = \mathcal{L}{cls} + \mathcal{L}{att}$;
    \STATE Update $\theta$, $\bm{W}$ and $\phi$ by minimizing $\mathcal{L}{pre}$;
    \ENDIF
    \STATE $\slash *$  Generative model training  $* \slash$
    \STATE Generate features $\{\tilde{\boldsymbol{X}}_{fe, k}^s, \boldsymbol{Z}_{k-1}^s\}$ for seen categories of stage $k$ using the generative model $G$.
    \STATE Update $\boldsymbol{Z}_{k-1}^s$ to $\boldsymbol{Z}_{k}^s$  based on $\bm{A}_k$;
    \STATE $\mathcal{L} = \mathcal{L}_{\mathrm{wgan}} + \lambda_{att} \times \mathcal{L}_{\mathrm{{anti-att}}} + \lambda_{fe} \times \mathcal{L}_{\mathrm{anti-fe}}$;
    \STATE Update $G$ and $D$ by minimizing $\mathcal{L}$ based on $\{\tilde{\boldsymbol{X}}_{fe, k}^s, \boldsymbol{Z}_{k}^s\}$;
    \STATE $\slash *$  Generate and replay features  $* \slash$
    \STATE Generate features $\{\tilde{\boldsymbol{X}}_{fe}^u, \boldsymbol{Z}^u\}$ for unseen categories of stage $k$ using the generative model $G$;
    \STATE Generate features $\{\tilde{\boldsymbol{X}}_{fe}^s, \boldsymbol{Z}^s\}$ for all history seen categories of stage $k$ using $G$;
    \STATE $\slash *$  Diagnosis model training  $* \slash$
    \STATE Train $\bm{W}_{k}^{new}$ by minimizing $\mathcal{L}{seen}$ using $\{\tilde{\boldsymbol{X}}_{fe}^s, \boldsymbol{Z}^s\}$;
    \STATE $\bm{W}_{k} = [\bm{W}_{k-1} | \bm{W}^{new}_{k}]$;
    \STATE Initialize $\bm{W}^{tmp}_k$ based on $\bm{W}_{k}$;
    \STATE Update  $\bm{W}^{tmp}_k$ based on equations (\ref{equ:memory_update}), (\ref{equ:memory_update2}), (\ref{equ:W-ATT_update}) using the generated features $\{\tilde{\boldsymbol{X}}_{fe}^u, \boldsymbol{Z}^u\}$;
    \ENDFOR
  \end{algorithmic}
\end{algorithm}

The training process for the attribute increment task is outlined in Algorithm \ref{alg:knowledge}. In each learning stage, we leverage the generative model to generate features for all seen categories. The attribute labels of the generated features are then updated based on the new attribute matrix $\bm{A}_k$. Subsequently, we retrain the generative model using the generated features and their corresponding relabeled attribute labels. This allows the generative model to incorporate the newly added fault attributes. As the fault attribute continues to expand, the generative model becomes capable of capturing the connections and distinctions between both seen and unseen categories from multiple perspectives. This results in improved generation performance for unseen categories. Additionally, we utilize the updated generative model to generate features for historical seen categories. These features are used to train the parts of the diagnosis model that correspond to the new fault attributes. After that, we generate features for unseen categories to update the attribute prototype matrix to enhance the diagnostic capability for unseen faults. In the model test process, the fault category of test samples is also determined by Equ. (\ref{equ:predict}).

\section{Experiments}
\label{experiments}
In this section, we present two case studies to evaluate the performance of the proposed method. The first case study focuses on a real hydraulic system and demonstrates the effectiveness of the proposed approach in terms of anti-forgetting and learning performance for category increment in generalized zero-shot fault diagnosis. The second case study is conducted on the Tennessee-Eastman process, a widely used benchmark for traditional zero-shot fault diagnosis. In this case, we showcase the learning capability of the proposed method when only new fault attributes are available.

\subsection{Hydraulic System}
\subsubsection{Introduction for Dataset}
The hydraulic system \cite{helwig2015condition} comprises a primary working circuit and a secondary cooling-filtration circuit, which are interconnected by an oil tank. The working circuit involves a main pump that operates with cyclically repeated load levels, controlled by a proportional pressure relief valve. Process sensors are employed to measure various parameters such as pressures, volume flows, temperature, electrical power, and vibration. Additionally, three virtual sensors are utilized to calculate cooling efficiency, cooling power, and system efficiency based on the readings from physical sensors. The sampling rates of the hydraulic sensors range from 100 Hz to 1 Hz. To ensure consistency in the sampling rates of the hydraulic sensors, a flattening technique is applied to the data\cite{zhuo2021auxiliary}. This process results in an original dataset consisting of 2205 instances and 43680 dimensions. The dataset encompasses a total of 144 fault categories ($3 \times 4 \times 3 \times 4$). These fault categories are described using four attributes: cooler condition (with 3 possible values), valve condition (with 4 possible values), internal pump leakage (with 3 possible values), and hydraulic accumulator (with 4 possible values). The distinct fault categories are determined based on different combinations of attribute values.

\subsubsection{Evaluation Metrics}
During the incremental learning process for the new seen classes, the objective of IZSFD is twofold: to preserve the diagnostic capability for the historical seen classes and to enhance the diagnostic capability for the new seen classes. To assess the diagnostic performance of the proposed method, we utilize the following evaluation metrics:
\begin{enumerate}
  \item[(a)] $acc_s$: The average accuracy of seen category faults.
    \begin{equation}
      acc_s = \frac{N_c^s}{N^s}
    \end{equation}
    where $N_c^s$ is the number of correctly classified samples and $N^s$ is the total number of seen fault samples.
  \item[(b)] $acc_u$: The average accuracy of unseen category faults.
    \begin{equation}
      acc_s = \frac{N_c^u}{N^u}
    \end{equation}
    where $N_c^u$ is the number of correctly classified samples and $N^u$ is the total number of unseen fault samples.
  \item[(c)] $Har$: The harmonic mean of accuracy in seen and unseen faults for evaluating the performance of GZSD.
    \begin{equation}
      Har = \frac{2 \times acc_s \times acc_u}{acc_s + acc_u}
    \end{equation}
\end{enumerate}
\subsubsection{Comparison methods}
Since there is no previous work for the category increment ZSFD, to evaluate the increment learning ability of the proposed method, we compare our proposed method with several classical increment learning strategies using the same network structure, but without the anti-forgetting designs proposed in our approach to conduct the comparison experiments, and we conduct a performance upper bound. The comparison methods are as follows:
\begin{enumerate}
  \item[(a)] \textit{Joint Learning (JL)}: The model is trained directly using all training sets from the first learning stage to the current learning stage, serving as an upper bound.
  \item[(b)] \textit{Replay based on Generation (RG)}: The model is trained using the training samples of the current learning stage and the generated samples of historical learning stages from the generative model of the previous learning stage.
  \item[(c)] \textit{Sequential Fine-Tuning (SFT)}: The model is fine-tuned sequentially with parameters initialized from the model fine-tuned in the previous learning stage.
  \item[(d)] \textit{L1 Regularization (L1)}: In each learning stage $k$, the generative model ($G_k$ and $D_k$) and the diagnosis model ($B_k$) are initialized as $G_{k-1}$, $D_{k-1}$, and $B_{k-1}$, and subsequently trained with L1 regularization applied between $G_k$ and $G_{k-1}$, $D_k$ and $D_{k-1}$, and $B_k$ and $B_{k-1}$.
  \item[(e)] \textit{L2 Regularization (L2)}: In each learning stage $k$, the generative model ($G_k$ and $D_k$) and the diagnosis model ($B_k$) are initialized as $G_{k-1}$, $D_{k-1}$, and $B_{k-1}$, and subsequently trained with L2 regularization applied between $G_k$ and $G_{k-1}$, $D_k$ and $D_{k-1}$, and $B_k$ and $B_{k-1}$.
  \item[(f)] \textit{Elastic Weight Consolidation (EWC)} \cite{2017Overcoming}: This method aims to keep the network parameters close to the optimal parameters from the previous stage while training the model for the current stage.
  \item[(g)] \textit{Memory Aware Synapses (MAS)} \cite{aljundi2018memory}: This method aims to accumulate an importance measure for each network parameter based on how sensitive the predicted output function is to a change in this parameter.
\end{enumerate}

\subsubsection{Implementation Details}
Our approach consists of a generator, a discriminator, and a diagnosis model. Both the generator and the discriminator are implemented as three linear layers and ReLU activation. The generator's hidden layer has 128 and 512 hidden units, while the discriminator's hidden layer has 512 and 32 units. The feature extractor of the diagnosis model is also a neural network with two linear layers and ReLU activation, comprising 256 hidden units. The attribute prototype matrix of the diagnosis model is a parameter matrix that can be learned during training. The learning rate is set to 0.0002, and the batch size of 16 is used. Each training step is conducted for 300 epochs.
The assignment of fault categories to different stages is done randomly. We perform five experiments and report the average results. The hyperparameters $\lambda$, $\lambda_{att}$, and $\lambda_{fe}$ are set to 10.
The threshold $\alpha_{limit}$ is set to 1.

\subsubsection{Results of Increment Zero-Shot Fault Diagnosis}
Table \ref{tab:category} presents the results of category increment generalized zero-shot fault diagnosis, indicating the average accuracy of seen and unseen faults across all learning stages. The Joint Learning (JL) method serves as an upper bound for performance since it utilizes all training samples from the first learning stage to the current stage. For seen faults, the proposed BDMAFF method consistently achieves the highest performance across all stages. This demonstrates the method's effectiveness in maintaining diagnostic capability for previously encountered faults. Additionally, the proposed method showcases superior performance in diagnosing unseen faults throughout all learning stages, indicating its proficiency in accumulating knowledge from previously encountered fault categories to improve generalization ability. The promising performance of the IZSFD method is expected to yield satisfactory results in terms of the $Har$ metric, which plays a crucial role in balancing the recognition of seen and unseen faults. Our method achieves the highest improvement of 4.46\% in terms of the $Har$ metric compared to all comparison methods.
\begin{table}[!t]
  \centering
  \caption{The average accuracy of seen and unseen categories for the category increment task}
  \label{tab:category}
  \begin{tabular}{cccc}
    \toprule
    \multirow{2}{*}{Method} & \multicolumn{3}{c}{Accuracy}                                   \\
    \cmidrule{2-4}
                            & Seen (\%)                    & Unseen (\%)    & Har (\%)       \\
    \midrule
    JL                      & 80.04                        & 59.29          & 68.12          \\
    RG                      & 69.36                        & 55.32          & 61.55          \\
    SFT                     & 69.80                        & 54.79          & 61.39          \\
    L1                      & 71.71                        & 55.26          & 62.42          \\
    L2                      & 69.79                        & 55.21          & 61.65          \\
    EWC                     & 68.53                        & 55.70          & 61.45          \\
    MAS                     & 74.30                        & 55.23          & 63.36          \\
    BDMAFF                  & \textbf{76.19}               & \textbf{57.99} & \textbf{65.85} \\
    \bottomrule
  \end{tabular}
\end{table}

\begin{figure}[!t]
  \centering
  \includegraphics[width=0.85 \linewidth]{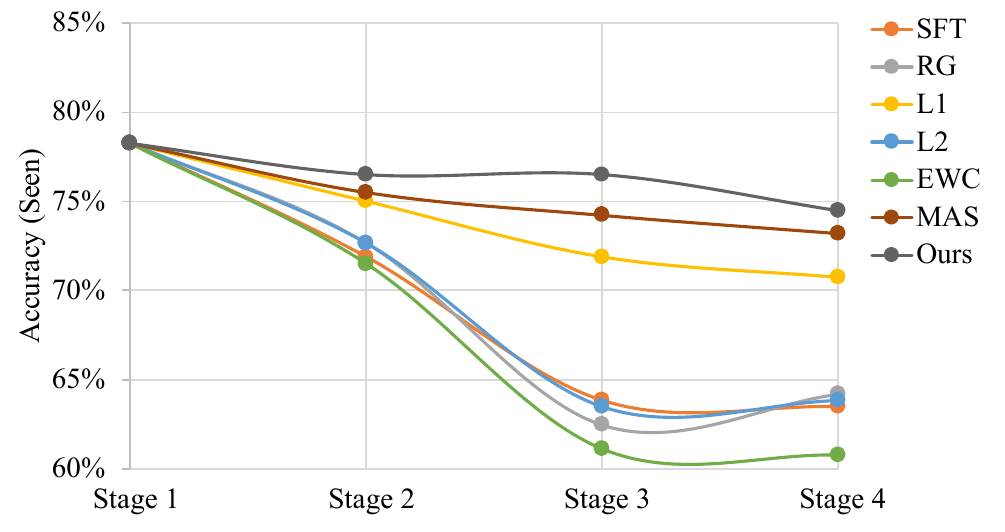}
  \caption{The accuracy of seen faults from Stage 1 in different learning stages.}
  \label{fig:seen_1}
\end{figure}

\begin{figure}[!t]
  \centering
  \includegraphics[width=0.85 \linewidth]{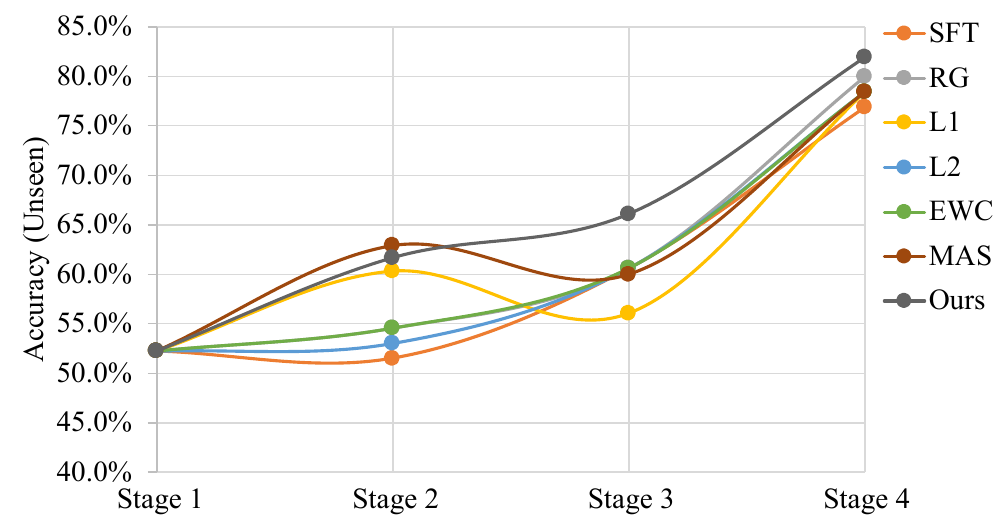}
  \caption{The accuracy of unseen faults of Stage 1 in different stages}
  \label{fig:unseen_1}
\end{figure}

To further illustrate the resistance to forgetting exhibited by each method, we evaluate the models trained at each learning stage using the Stage 1 test set, employing the class-attribute matrix used in Stage 1. As depicted in Fig. \ref{fig:seen_1}, a decline in diagnostic accuracy for seen categories is observed as the learning stages progress, indicating the presence of forgetting. As the model learns new faults, its diagnostic capability for previously seen categories diminishes. However, in comparison to other methods, our approach consistently achieves the highest diagnostic accuracy throughout all stages, demonstrating its effectiveness in mitigating forgetting. This can be attributed to our proposed BDMAFF, which leverages the memory of historical fault features and incorporates a memory matrix to preserve the memory of attribute prototypes. Moreover, it can be observed that, for certain methods, the diagnostic performance for Stage 1 seen faults in later learning stages occasionally surpasses that of the previous stage. This can be attributed to increased exposure to seen fault categories during training, leading to improved generalization of the model. In other words, as the learning stages progress, the diagnostic capability for historical seen categories decreases due to forgetting, but the emergence of new faults may contain similar information to historical faults, thus aiding in better diagnosing historical faults. Regarding unseen faults, as depicted in Fig. \ref{fig:unseen_1}, our method achieves the highest accuracy across most of the stages, showcasing its ability to effectively integrate information learned in different stages and enhance generalization for unseen faults during the model's updating. 
It is worth noting that since Task I has the fewest categories and involves a relatively simpler classification task, the accuracy appears higher than the average accuracy of all tasks presented in Table \ref{tab:category}.

\begin{figure}[!t]
  \centering
  \captionsetup[subfloat]{font=scriptsize}
  \vspace{-20pt}
  \subfloat[]{\label{fig:subfig:a}\includegraphics[scale=0.16]{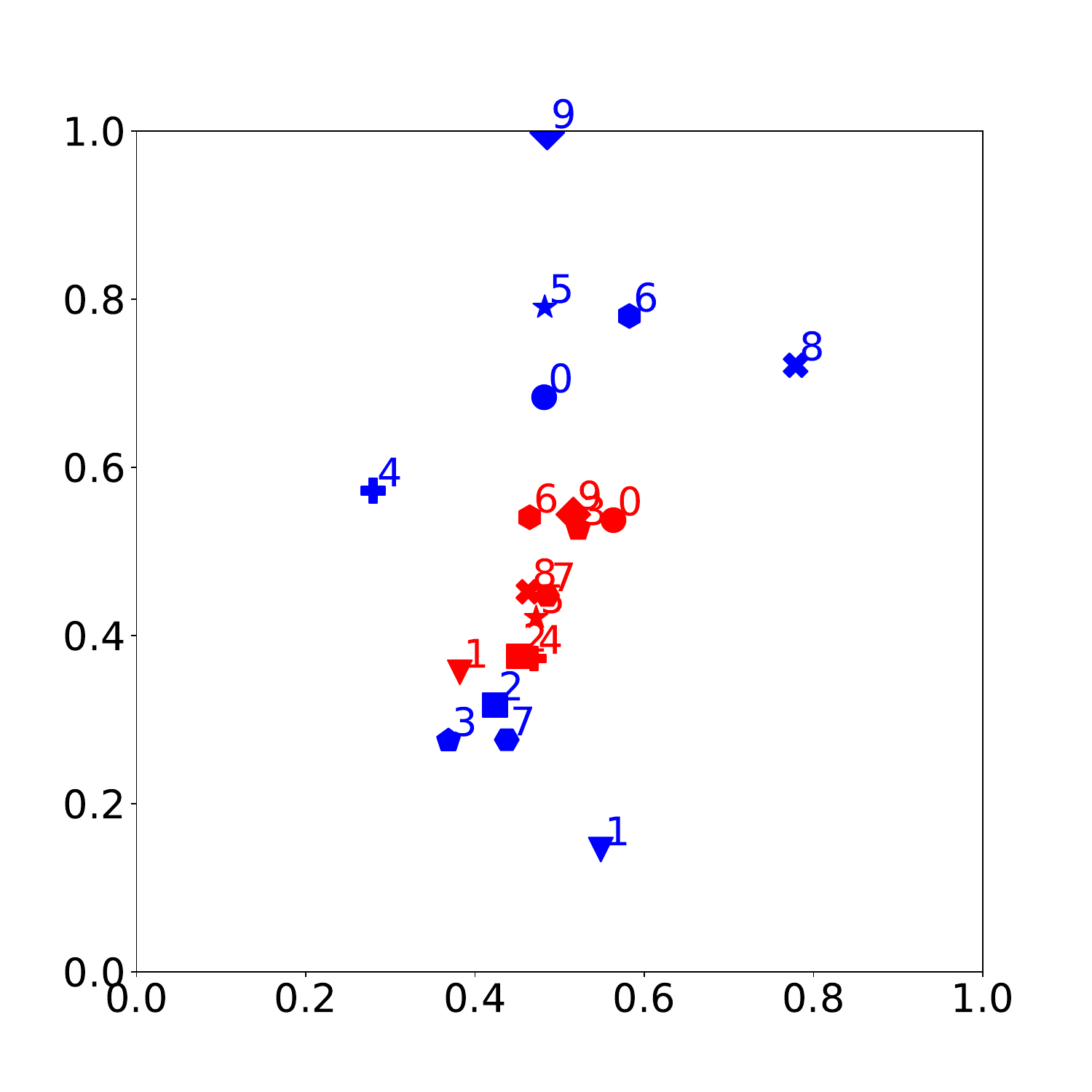}}
  \subfloat[]{\label{fig:subfig:b}\includegraphics[scale=0.16]{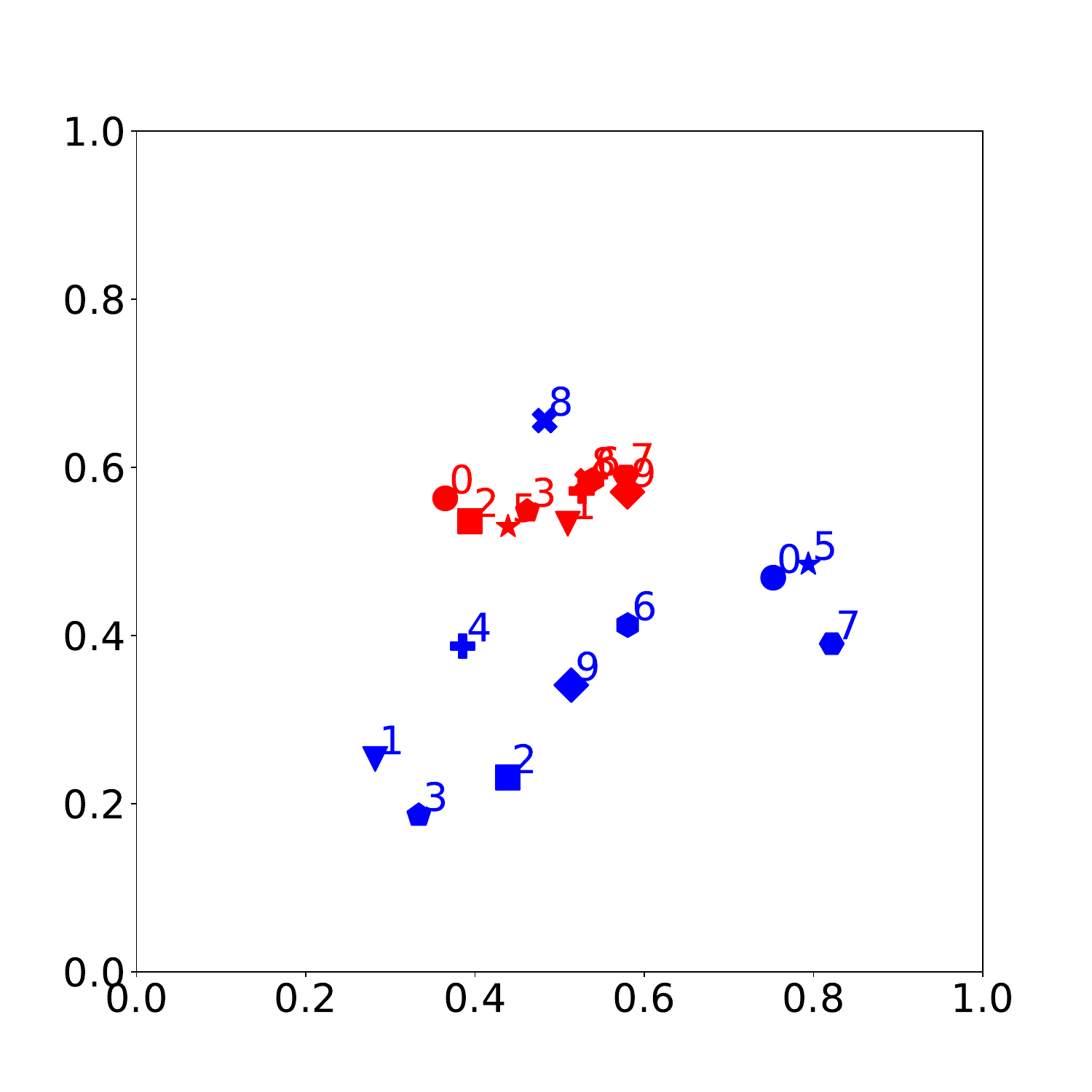}} \\
  \vspace{-11pt}
  \subfloat[]{\label{fig:subfig:c}\includegraphics[scale=0.16]{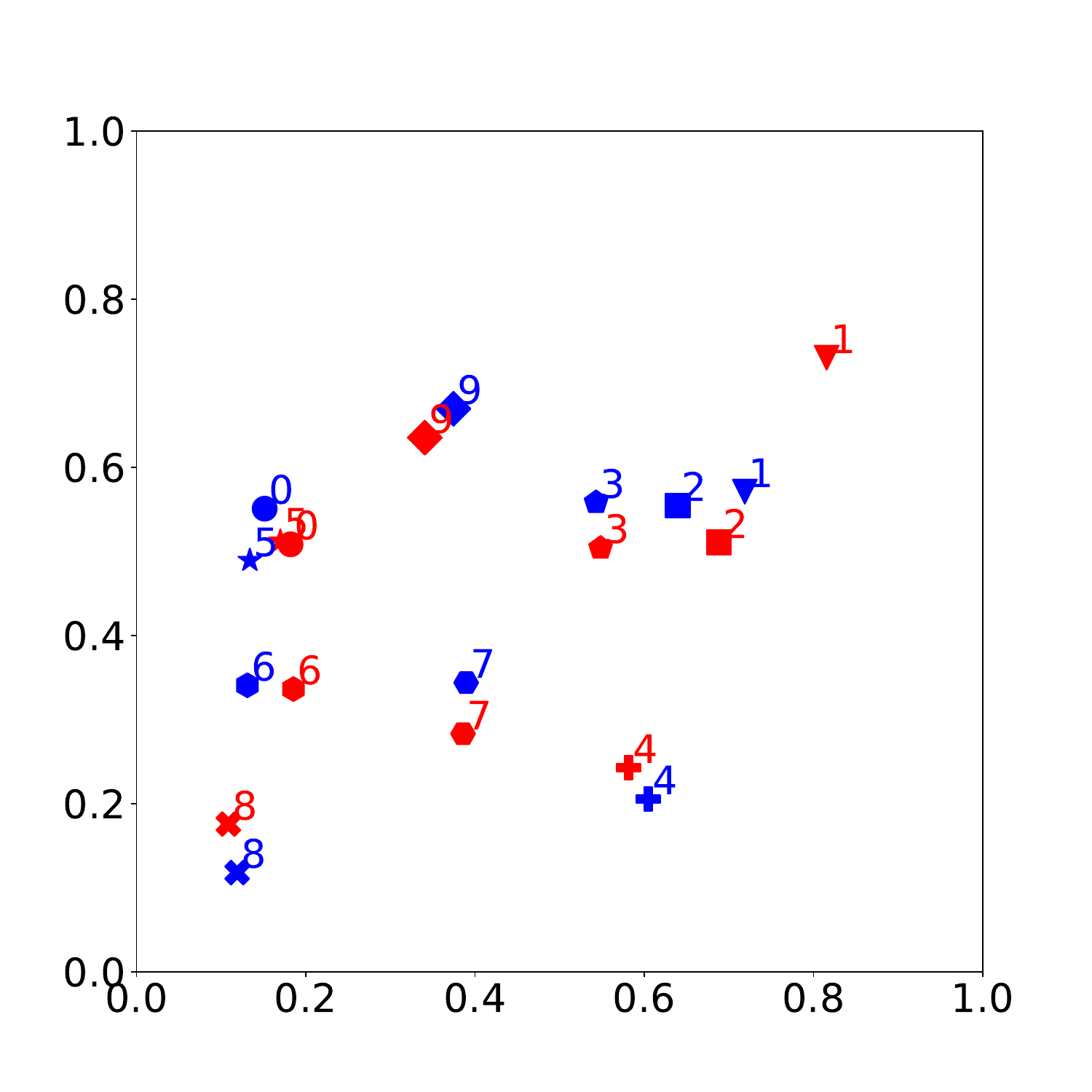}}
  \subfloat[]{\label{fig:subfig:d}\includegraphics[scale=0.16]{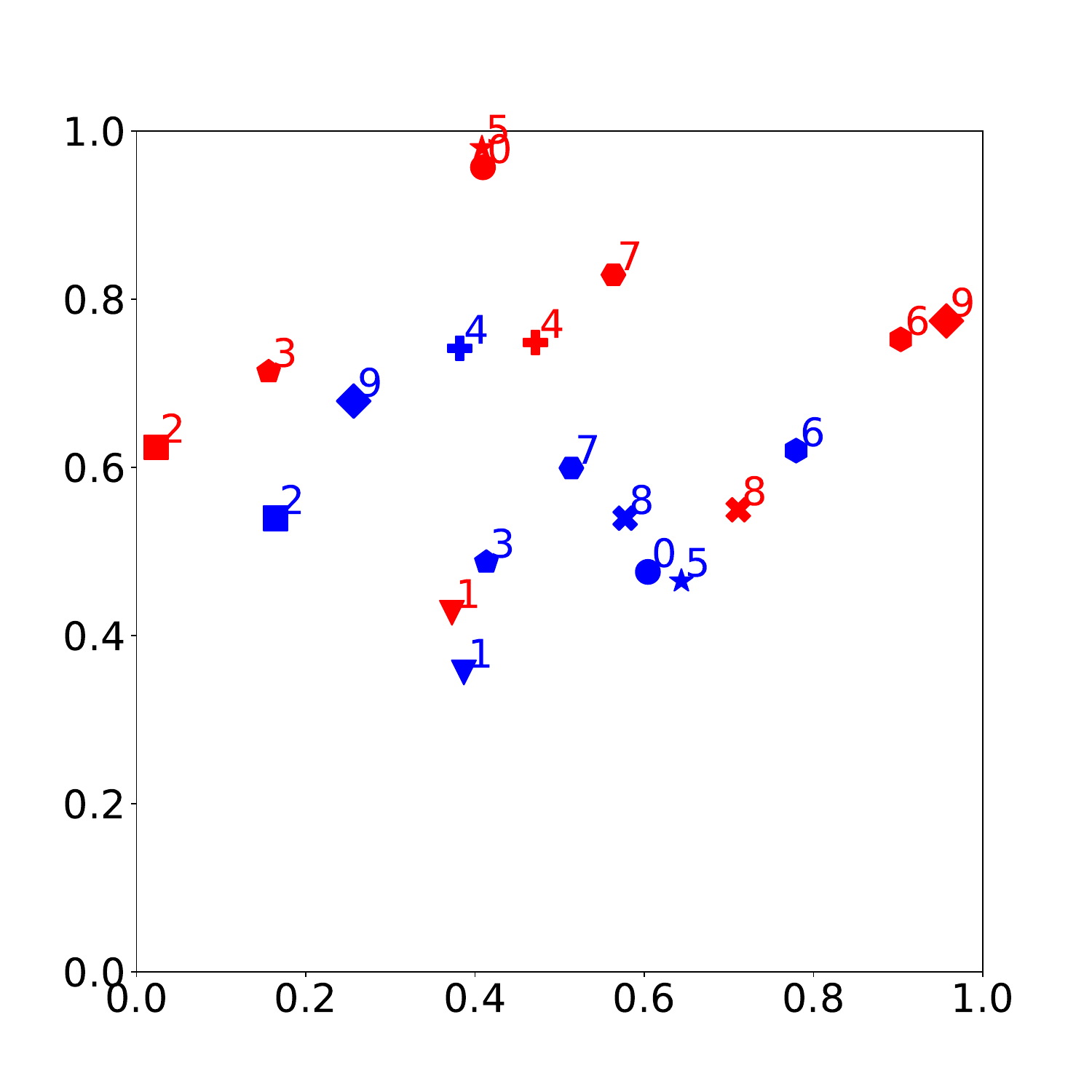}} \\
  \vspace{-11pt}
  \subfloat[]{\label{fig:subfig:e}\includegraphics[scale=0.16]{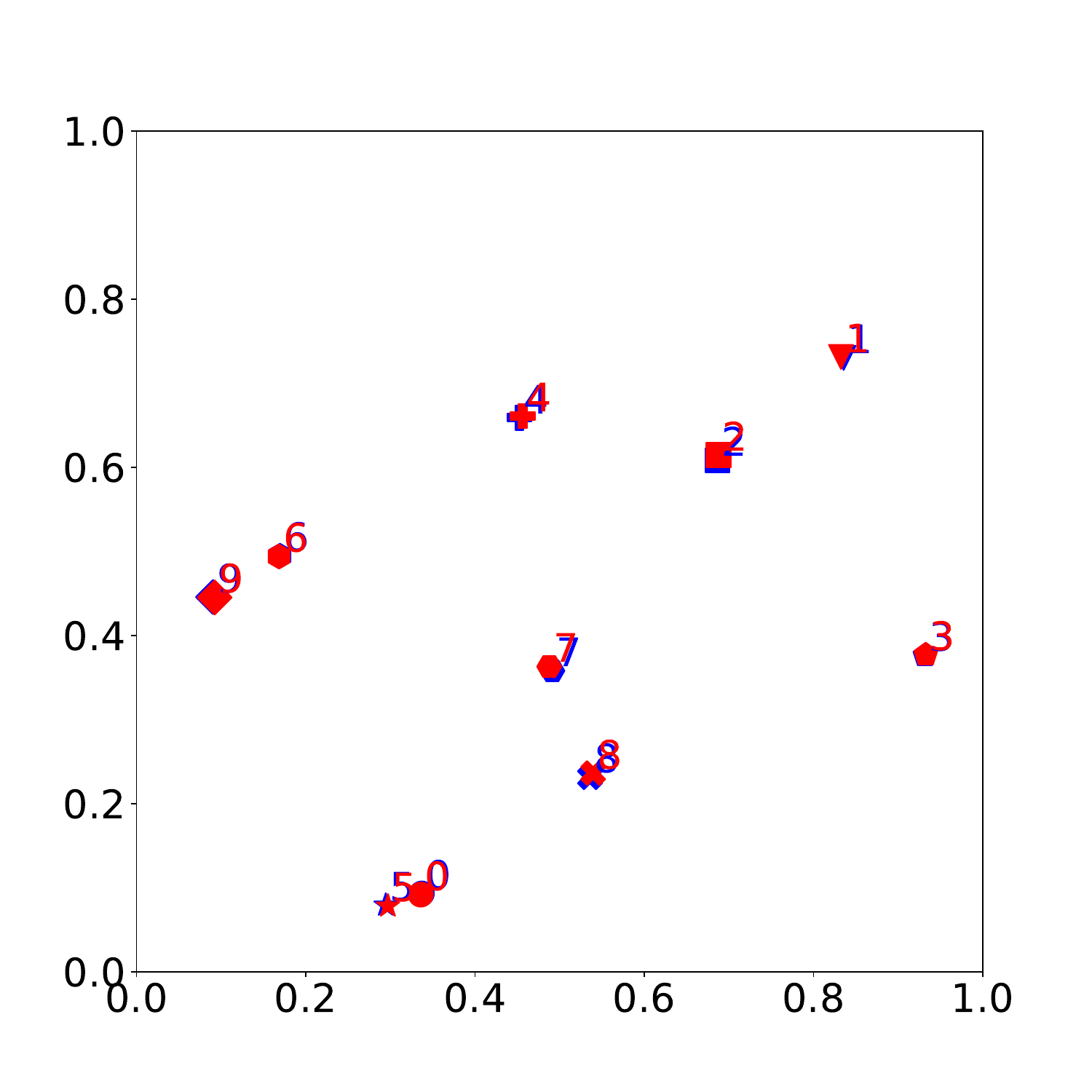}}
  \subfloat[]{\label{fig:subfig:f}\includegraphics[scale=0.16]{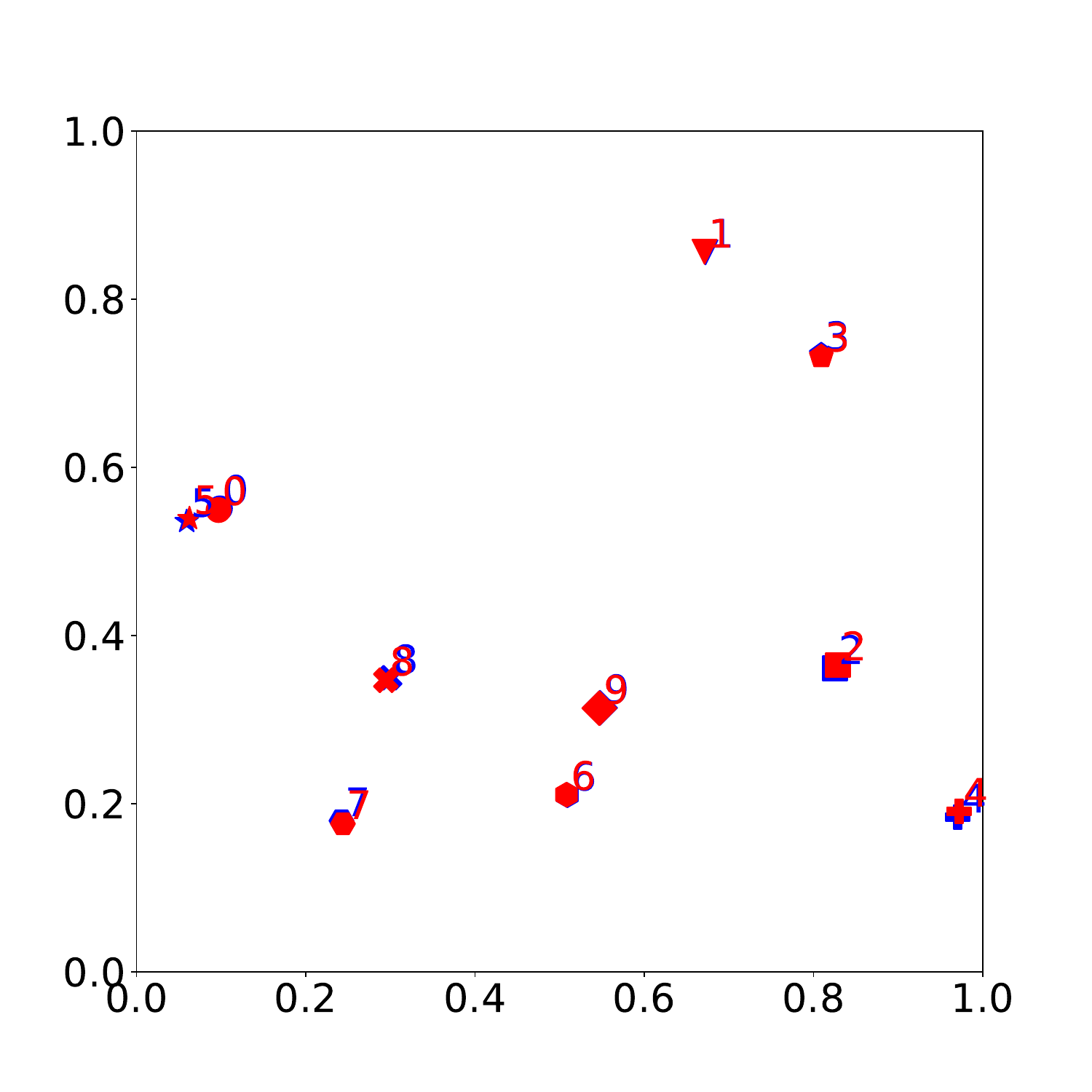}}
  \caption{The centers of generated features for the hydraulic system. (a) and (b) depict the generated features of Stage 1 and Stage 4, respectively, using the generative model trained with $\mathcal{L}_{\mathrm{wgan}}$. (c) and (d) represent the generated features of Stage 1 and Stage 4, respectively, using the generative model trained with $\mathcal{L}_{\mathrm{wgan}}$ and $\mathcal{L}_{\mathrm{\mathrm{anti-att}}}$. (e) and (f) display the generated features of Stage 1 and Stage 4, respectively, using the generative model trained with $\mathcal{L}_{\mathrm{wgan}}$, $\mathcal{L}_{\mathrm{\mathrm{anti-att}}}$, and $\mathcal{L}_{\mathrm{\mathrm{anti-fe}}}$.}
  \label{fig:generative}
\end{figure}

\subsubsection{Analysis of Anti-Forgetting Strategy for the Generative Model}
To assess the effectiveness of our proposed anti-forgetting strategy for the generative model, we visualize and compare the generated features with real features from both Stage 1 and Stage 4. We select the top ten seen classes from Stage 1 for visualization. Our experiments involve comparing the generated features of the generative model trained solely with the WGAN loss ($\mathcal{L}_{\mathrm{wgan}}$), the introduced attribute anti-forgetting loss ($\mathcal{L}_{\mathrm{\mathrm{anti-att}}}$), and the introduced feature anti-forgetting loss ($\mathcal{L}_{\mathrm{\mathrm{anti-fe}}}$). From Fig. \ref{fig:subfig:a} and \ref{fig:subfig:b}, it is evident that the generated features of the generative model exhibit poor performance, indicating the difficulty of accurately generating features. Furthermore, by comparing Fig. \ref{fig:subfig:a} and Fig. \ref{fig:subfig:b}, we observe that as the training stage progresses, the generated samples deviate further from the real samples due to forgetting issues. When the attribute anti-forgetting loss ($\mathcal{L}_{\mathrm{\mathrm{anti-att}}}$) is introduced, as shown in Fig. \ref{fig:subfig:c}, the generated features become closer to the real features by incorporating the diagnosis model to supervise the quality of the generated samples. However, a significant impact of forgetting issues is still apparent when comparing Fig. \ref{fig:subfig:c} and Fig. \ref{fig:subfig:d}. Upon further introduction of the feature anti-forgetting loss ($\mathcal{L}_{\mathrm{\mathrm{anti-fe}}}$), as depicted in Fig. \ref{fig:subfig:e}, the generated features align effectively with the real features. This demonstrates the successful memorization of information from the seen classes by the generative model. Moreover, by comparing Fig. \ref{fig:subfig:e} and Fig. \ref{fig:subfig:f}, it is evident that the forgetting issue is significantly alleviated, as the model retains its memory capacity for the seen classes from the first stage even after multiple learning stages.

\subsection{Tennessee-Eastman Process}
\subsubsection{Introduction for Dataset}
\begin{figure}[!t]
  \centering
  \includegraphics[width=\linewidth]{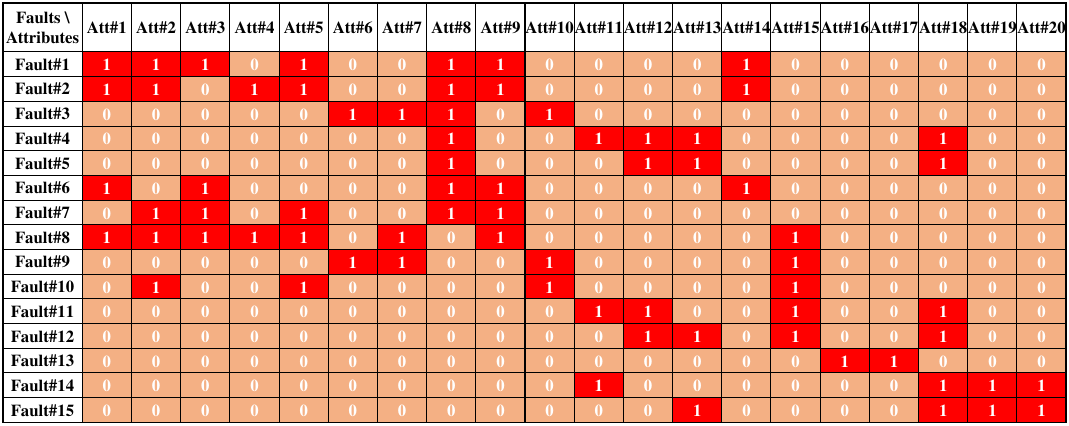}
  \caption{Fault semantic attribute description. The "1" in the figure denotes the fault has this attribute, and the "0" denotes not.}
  \label{fig:ATTmatrix}
\end{figure}

In this section, we investigate the Tennessee-Eastman process (TEP) \cite{downs1993plant}, a widely used dataset for evaluating the performance of traditional ZSFD methods \cite{zhuo2021auxiliary, 9904860, feng2020fault}. The TEP dataset comprises five major subsystems: reactor, condenser, vapor-liquid separator, recycle compressor, and product stripper. It consists of 41 measured variables and 11 manipulated variables, along with 21 fault types in addition to the normal state. We select 15 fault types for training and testing.
The fault attributes for the TEP dataset are shown in Fig. \ref{fig:ATTmatrix}.
We randomly divide the seen and unseen faults and conduct five experiments, reporting the average results. For each fault, we gather 480 samples for training and 800 samples for testing. Therefore, the total number of training samples is 12*480, while the total number of test samples is 15*800 (including 12*800 samples for seen faults and 3*800 samples for unseen faults).

\subsubsection{Comparison methods}
Regarding zero-shot fault diagnosis with attribute increment, to the best of our knowledge, no previous works have specifically addressed this problem. Existing incremental learning algorithms have been designed for data increment rather than attribute increment, as explained in Section \ref{sec:challenges}. Therefore, to enable a comparison with our method, the existing methods are trained in a joint learning manner, where the comparison models are trained using all the training samples and their attribute labels at each learning stage. As the learning stages progress, the number of fault attributes increases, increasing the dimensionality of the attribute labels. In contrast, our method adopts an attribute increment paradigm, where the training data is only accessible in the first learning stage. In the subsequent learning stages, only the newly added fault class-attribute matrix is provided. It is evident that attribute increment poses greater challenges compared to the joint learning manner, and the experimental results of the comparison methods can be considered as performance upper bounds to some extent. Additionally, we also provide the results of our method trained in a joint learning manner. We adopt FDAT\cite{feng2020fault} and SCE\cite{9904860} for comparison.

\subsubsection{Results of Increment Zero-Shot Fault Diagnosis}
The results of traditional zero-shot fault diagnosis with attribute increment are presented in Table \ref{tab:knowledge}. As the learning stages progress, all methods show improved diagnostic accuracy for unseen categories, indicating the crucial role of fault attributes in transferring diagnostic capability from seen to unseen categories. The accumulation of fault attributes leads to enhanced generalization performance for unseen fault categories within the traditional zero-shot fault diagnosis methods. In the case of our proposed method, even in the attribute increment setting where historical training samples are unavailable in subsequent learning stages and only the updated fault class-attribute matrix is provided, it still achieves comparable performance to the comparison methods. This demonstrates the effectiveness of our approach in preventing forgetting and accumulating historical knowledge.

\begin{table}[!t]
  \centering
  \caption{The average accuracy of unseen categories for the attribute increment task}
  \begin{tabular}{ccccc}
    \toprule
    Method                        & Classifier   & Stage 1 & Stage 2 & Stage 3 \\
    \midrule
    \multirow{3}[2]{*}{FDAT (JL)} & LSVM         & 57.35\% & 63.42\% & 68.77\% \\
                                  & NRF          & 54.21\% & 57.23\% & 71.13\% \\
                                  & PNB          & 37.48\% & 56.23\% & 59.15\% \\
    \midrule
    \multirow{3}[2]{*}{SCE (JL)}  & LSVM         & 68.50\% & 68.77\% & 76.26\% \\
                                  & NRF          & 65.48\% & 70.07\% & 72.35\% \\
                                  & PNB          & 55.40\% & 67.94\% & 72.27\% \\
    \midrule
    BDMAFF                        & $\backslash$ & 65.99\% & 68.69\% & 72.49\% \\
    BDMAFF (JL)                   & $\backslash$ & 65.99\% & 71.30\% & 76.89\% \\
    \bottomrule
  \end{tabular}%
  \label{tab:knowledge}%
\end{table}%

\section{Conclusion}
\label{conclusion}
This work is the first attempt to address the category-attribute increment issue of the existing zero-shot fault diagnosis paradigms, which refers to the ability to learn new fault categories and attributes while retaining knowledge learned previously. To deal with this issue, we propose the incremental zero-shot fault diagnosis paradigm, which involves both category and attribute increment tasks for traditional and generalized zero-shot fault diagnosis scenarios.
A broad-deep mixed anti-forgetting framework is designed, which mitigates forgetting historical seen faults and enhances the diagnostic capability for unseen faults when learning new categories or attributes.
It addresses the issue of forgetting in increment tasks from the perspectives of feature memory and attribute memory.
On the one hand, the deep generative model retains the memory of features and transfers accumulated knowledge to the diagnosis model by generating features for unseen faults.
The forgetting-resistant loss functions from both attribute and feature perspectives mitigate the forgetting issue of the generative model.
On the other hand, the diagnosis model is equipped with a memory matrix, which retains attribute prototypes' memory for anti-forgetting with the help of the generative model.
A memory-driven iterative update strategy is designed, which can update these prototypes without storing samples of historical stages.
The effectiveness of the proposed method is evaluated through two case studies involving a real hydraulic system and the Tennessee-Eastman process.
For the category increment task, our method achieves the highest improvement of 4.46\% in terms of the harmonic mean of seen and unseen faults compared to all the baseline methods. Additionally, through ablation experiments, the effectiveness of the proposed anti-forgetting losses is demonstrated in mitigating the forgetting issue of the generative model.
For the attribute increment task, in the new learning stage, our method achieves competitive results compared to the baseline methods, even without using real training samples, while the baseline methods rely on stored real training samples.

\bibliographystyle{IEEEtran}  
\bibliography{IEEEabrv, tii-articles-template.bib}

\begin{IEEEbiography}[{\includegraphics[width=1in,height=1.25in,clip,keepaspectratio]{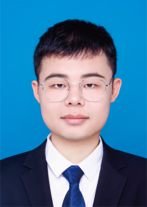}}]{Jiancheng Zhao} received B.Eng. degree in automation from College of Control Science and Engineering, Zhejiang University, Hangzhou, China, in 2021, where he is currently pursuing the Ph.D. degree. His current research interests include industrial big data, zero-shot learning.
\end{IEEEbiography}

\begin{IEEEbiography}[{\includegraphics[width=1in,height=1.25in,clip,keepaspectratio]{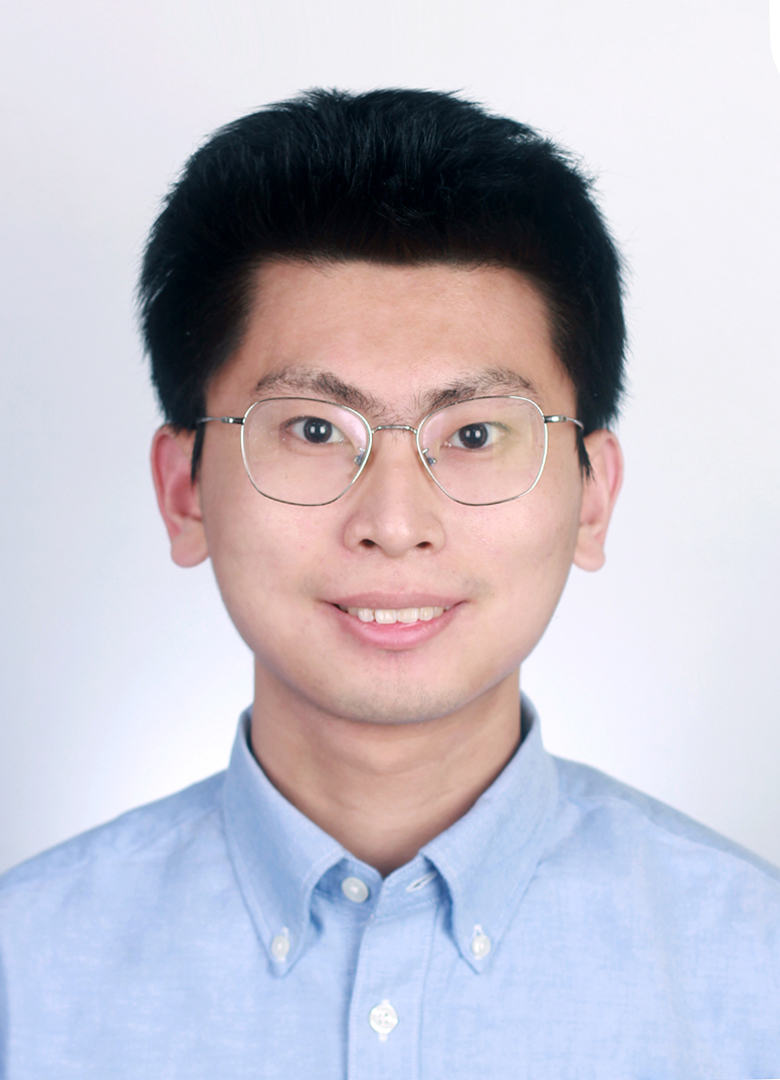}}]{Jiaqi Yue} received the B.Eng. degree in automation from the College of Electrical Engineering, Zhejiang University, Hangzhou, China, in 2022, where he is currently pursuing the Ph.D. degree in control science and engineering with the College of Control Science and Engineering. His current research interests include fault diagnosis and zero-shot learning.
\end{IEEEbiography}

\begin{IEEEbiography}[{\includegraphics[width=1in,height=1.25in,clip,keepaspectratio]{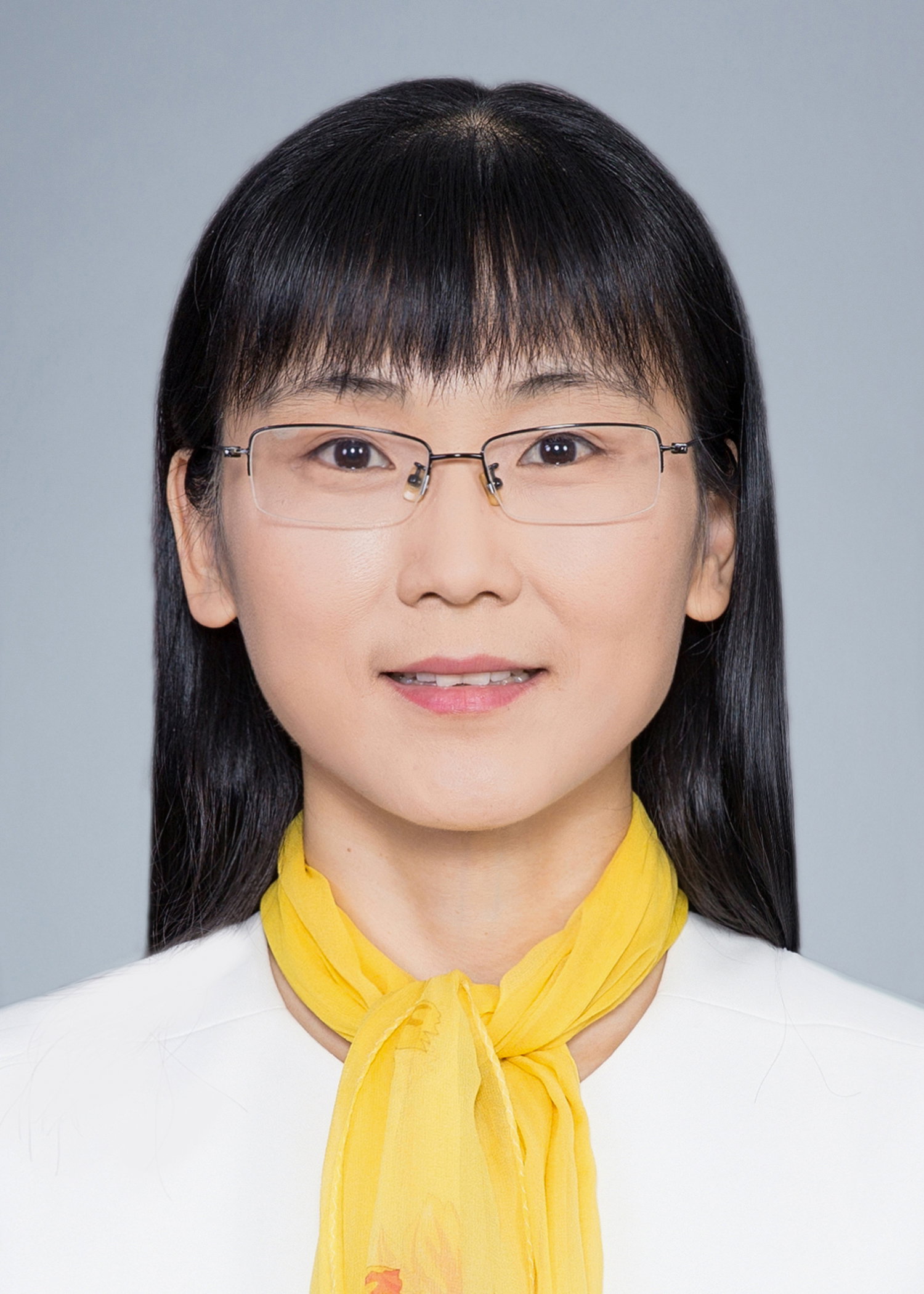}}]{Chunhui Zhao}
  (SM'15) received Ph.D. degree from Northeastern University, China, in 2009. From 2009 to 2012, she was a Postdoctoral Fellow with the Hong Kong University of Science and Technology and the University of California, Santa Barbara, Los Angeles, CA, USA. Since January 2012, she has been a Professor with the College of Control Science and Engineering, Zhejiang University, Hangzhou, China. Her research interests include statistical machine learning and data mining for industrial applications. She has authored or co-authored more than 140 papers in peer-reviewed international journals. She has served Senior Editor of Journal of Process Control, AEs of two International Journals, including Control Engineering Practice and Neurocomputing.

  She was the recipient of the National Top 100 Excellent Doctor Thesis Nomination Award, New Century Excellent Talents in University, China, and the National Science Fund for Excellent Young Scholars, respectively.
\end{IEEEbiography}

\vfill

\end{document}